\documentclass{article}

\usepackage{PRIMEarxiv}

\usepackage[utf8]{inputenc} 
\usepackage[T1]{fontenc}    
\usepackage{hyperref}       
\usepackage{url}            
\usepackage{booktabs}       
\usepackage{amsfonts}       
\usepackage{nicefrac}       
\usepackage{microtype}      
\usepackage{lipsum}
\usepackage{fancyhdr}       
\usepackage{graphicx}       
\graphicspath{{media/}}     
\usepackage{listings}
\usepackage{booktabs}
\usepackage{float}
\usepackage{hyperref}
\usepackage{orcidlink}
\usepackage{fontawesome5}

\title{PromptMN: Pseudo Prompting Language
}

\author{
  Enkhzol Dovdon\,\orcidlink{0000-0001-6952-8960}\\
  ICT Group\\
 \texttt{enkhzol.dovdon@ict.nl} \\
}

\begin{document}
\maketitle

\begingroup
\renewcommand{\thefootnote}{}
\footnotetext{
\faGithub\ Code: \href{https://github.com/denkhzol/PromptMN}{\texttt{github.com/denkhzol/PromptMN}}
}
\endgroup

\begin{abstract}
Prompting has become the primary interface between humans and generative AI, yet many natural language prompts remain fragile: roles, goals, constraints, and expected outputs are often buried in prose or left implicit. In agentic and software development workflows, a misread at the first handoff can propagate through every step, since a significant portion of agent failures stem from context ambiguities rather than model limitations. This paper introduces PromptMN, a pseudo-prompting domain-specific language that annotates natural language with compact, \%-prefixed typed directives covering roles, goals, requirements, priorities, constraints, plans, inputs, and outputs. Semantic resolution lets authors write in any order while the model interprets directives by function. PromptMN sits between informal prompting and programming-style pseudocode: structured enough to be inspectable and reusable, yet lightweight enough for analysts, managers, developers, and stakeholders across the software development lifecycle (SDLC). PromptMN also pairs with reverse prompt engineering. Asking a model to restate a desired outcome as PromptMN lets users inspect the inferred roles, goals, constraints, and missing assumptions before acting, reducing repair cycles and yielding a reusable artifact for aligning people and AI tools. PromptMN's feasibility is evaluated across several frontier models, including Claude Fable 5, Claude Opus 4.8, Gemini 3.1 Pro, and GPT-5.5. The models correctly resolved PromptMN instructions, including complex structures such as repetition, conditionals, methods, and a prime-checking task, without fine-tuning. The same vocabulary applies across new codebases, maintenance, and redesign in the SDLC scenarios presented. While large-scale validation remains future work, these early results suggest PromptMN is a practical step toward clearer, more reviewable human-to-AI interaction.

\end{abstract}



\section{Introduction}
\label{intro}

The telephone game is a familiar version of what psychology calls serial reproduction, where a message changes as it passes through a chain of interpretation~\cite{bartlett1995remembering}. In that setting, the distortion occurs between humans and grows with every retelling. In the AI era, generative AI has made human-to-AI interaction a much more common part of Human-Computer Interaction (HCI), with people asking Large Language Models (LLMs) and agentic systems to infer goals, constraints, priorities, and acceptable outputs from ordinary language. The first handoff is where this distortion enters: if intent is not conveyed clearly at the start of the interaction, the model commits to a flawed reading, and agentic systems can carry it forward and amplify it at each step, much as later players repeat a message that has already drifted. In daily work and especially across the software development lifecycle, the quality of that initial inference often decides whether AI support saves time or creates rework.

LLMs are useful because they can interpret context, summarize information, generate structured content, write code, plan steps, and call tools. Recent models span billions to trillions of parameters, from 8B and 70B systems to larger 405B and near-trillion-parameter models. Parameter count does not directly equal stored knowledge, but it is a proxy for representational capacity: Morris et al.~\cite{morris2025much} estimate that GPT-family models hold an approximate capacity of 3.6 bits per parameter. By that measure, a 70B-parameter model may encode on the order of tens of gigabytes (roughly 30\,GB) of compressed, distributed knowledge. This knowledge is not exposed as a clear database that a user can inspect; it lives in model weights as statistical patterns and is retrieved probabilistically. The prompt, therefore, becomes the main interface for selecting and shaping what the model should do. Effective prompting is more than polite wording. It is the practical act of translating human intent into a form that a model can recognize, organize, and execute~\cite{ekin2026prompt, liu2026comprehensive}.

Current prompt practice still relies heavily on free-form prose. Guides commonly recommend roles, background context, examples, constraints, output formats, and iteration, and context engineering frames these choices as the work of supplying the right information at the right time~\cite{hua2025context, savva2026context}. Yet in real tasks these elements are often buried inside long paragraphs, repeated in slightly different forms, or left implicit. This is costly for beginners because they may not know which pieces matter, and it is also awkward for experienced software practitioners because their working mindset is usually shaped by explicit interfaces, requirements, types, and control structures rather than open-ended semantic guessing.

The needed shift is therefore not simply from short prompts to longer prompts. It is from treating a prompt as a chat message to treating it as an inspectable expression of intent. A prompt that drives analysis, coding, planning, or an agentic workflow should make its assumptions, requirements, boundaries, and requested outputs visible before work begins. This does not remove natural language; it gives natural language a small amount of structure so that both human and model can reason over the same task description.

PromptMN is introduced as a pseudo-prompting domain-specific language for that purpose. It uses compact reserved keywords, prefixed with `\%`, as typed semantic directives inside ordinary prose. Keywords such as `\%role`, `\%goal`, `\%req`, `\%should`, `\%could`, `\%mustnot`, `\%plan`, `\%showplan`, `\%in`, and `\%out` name the parts of a prompt that users already try to express, but often express inconsistently. Its requirement labels also reflect MoSCoW prioritization, where Must Have, Should Have, Could Have, and Won't Have categories guide requirements planning and often map the first three groups to 60\%, 20\%, and 20\% allocation heuristics~\cite{miranda2022moscow}. The language is intended to be mnemonic rather than heavy: a user can write a full PromptMN block for a complex task, or attach a few directives to a normal prompt when only lightweight structure is needed.

This design has three practical aims. First, it reduces wording effort by giving common prompt responsibilities stable names. Second, it works as a reminder template, helping authors include context, constraints, priorities, plans, and output expectations without repeatedly inventing a structure from scratch. Third, it makes prompts easier to inspect, reuse, and revise. In reverse prompting, for example, a model can restate a user's request as PromptMN so the user can see the model's inferred plan, assumptions, and missing constraints before execution. The result is a clearer negotiation between human intention and machine interpretation.

PromptMN sits between informal prompting and full pseudocode. It does not ask non-programmers to become software engineers, and it does not ask software engineers to abandon natural language. Instead, it offers a shared notation for business analysts, managers, developers, and AI-assisted development tools that already depend on structured intent. At a high level, PromptMN provides small semantic pieces that can be combined for a user's purpose, like assembling separate puzzle pieces into one task picture. The contribution of this paper is to define that notation, explain its design rationale, and show how it can support more reliable human-AI collaboration without replacing the flexibility that makes prompting useful in the first place. 

The remainder of this paper is organized as follows. Section~\ref{related} surveys prompt engineering, context engineering, and prior structured-prompt work, and positions PromptMN relative to both informal practice and programming-style pseudocode. Section~\ref{Methodology} defines the language, covering its syntax, keyword clusters, semantic resolution rule, and interpreter contract. Section~\ref{CaseStudies} demonstrates PromptMN recognition by current models through examples ranging from basic control flow to SDLC-oriented scenarios and reverse prompt engineering. Section~\ref{Conclusion} reflects on limitations and future work.

\section{Related Works}
\label{related}
Prompt engineering is now an established practice with a documented toolkit of techniques, such as role prompting, chain-of-thought, few-shot exemplars, and explicit constraints~\cite{ekin2026prompt, liu2026comprehensive, choi2025survey, maaz2025guide}. Liu et al.~\cite{liu2026comprehensive} sort these into four aspects, namely profile and instruction, knowledge, reasoning and planning, and reliability, and note that unstructured prompts tend to give variable, less reliable answers. Li et al.~\cite{li2025survey} tie the fragility of manual prompting to expert dependency and format sensitivity, where small shifts in punctuation or phrasing change the result. Beginner guides agree, since a good prompt has to set a role, limit the scope, and fix the output format~\cite{choi2025survey, ekin2026prompt}, yet they leave this as advice rather than enforceable structure. PromptMN recasts these levers as typed keywords like \texttt{\%role}, \texttt{\%mustnot}, and \texttt{\%format}, so the recommended structure becomes explicit and checkable.

Context engineering moves the focus from the model to the input it receives. Savva~\cite{savva2026context} describes context provision, that is problem framing, information selection, constraints, and evaluation criteria, as a costly input that hinges on expertise and is hard to observe, and argues that this context, not raw model capability, sets the value of the output. Hua et al.~\cite{hua2025context} define the practice as designing and organising context so machines act in line with human intent, and trace it over a twenty-year path from human-computer interaction to today's agents. The evidence backs this up, because Adam et al.~\cite{adam2026prompt} report that screening F1 swung widely with prompt composition and improved when prompts named the task and listed selection criteria. Gibreel et al.~\cite{gibreel2026prompt} read prompt-engineering competence as the skill of stating goals, constraints, and context clearly, while MEMENTO~\cite{kontonis2026memento} shows that a model's own reasoning benefits from being split into blocks and compressed. These findings motivate PromptMN's intent and requirement clusters, which reserve a fixed slot for framing, constraints, and success criteria.

The need for structure grows once prompts begin to drive autonomous, multi-step work. Practical guides cast an agent as a system that reasons, plans, and acts toward a goal with limited human involvement~\cite{aws2025agentic}, and production reports favour workflow decomposition, deterministic orchestration, and externalised prompts over one monolithic prompt~\cite{bandara2025practical}. Pattern catalogues and SDLC reviews repeat the same parts, such as prompt chaining, evaluator and reflect-refine loops, and role specialisation with iterative feedback~\cite{aws2026agentic, tang2026llm}. ReAct~\cite{yao2022react} adds that weaving reasoning traces through actions curbs hallucination and error propagation while keeping each step auditable. PromptMN moves these primitives into the prompt through ordered steps and plan disclosure (\texttt{\%plan}, \texttt{\%showplan}, \texttt{\%trace}), control flow (\texttt{\%if}, \texttt{\%repeat}, \texttt{\%method}), and explicit data boundaries (\texttt{\%in}, \texttt{\%out}).

The closest prior efforts are those that give prompts a code-like form. The Super-NaturalInstructions study~\cite{wang2022super} introduced a large collection of NLP tasks to examine how language models generalize from natural-language instructions. Each task combines a definition, positive and negative examples, an input, and an expected output. The results show that example-based instructions improve cross-task transfer, but they also reveal sensitivity to wording and a dependence on high-quality examples. PromptMN builds on this by transforming the structure into a pseudo-prompting language with explicit rules and constraints rather than free-form text.

Mishra et al.~\cite{mishra2023prompting} rewrote 132 Super-NaturalInstructions tasks into Python-style pseudocode prompts with typed functions, docstrings, and subtask calls. On BLOOM and CodeGen models (2B--7B), these prompts improved classification performance by 7--16 F1 points and ROUGE-L by 12--38\%, particularly for code-trained models. They also found that comments, docstrings, and structural elements matter, while writing pseudocode requires technical expertise, limiting its widespread usage. Similarly, Kumar et al.~\cite{kumar2025training} train LLMs to translate natural-language instructions into Python-like pseudocode before response generation. Fine-tuned on augmented TuluV2 data, their approach improves instruction-following performance by 8--21\% across 12 benchmarks without altering the user interaction paradigm, and in some cases even improves performance on mathematical and commonsense reasoning tasks.

Taken together, this body of work shows that structure, whether roles, constraints, plans, control flow, or traceable reasoning, makes prompts more reliable, yet it arrives either as informal advice or as full pseudocode that demands programming skill~\cite{mishra2023prompting, kumar2025training}. PromptMN sits between these, since it is a compact, mnemonic keyword set that annotates ordinary prose and resolves before execution, which keeps prompt structure reviewable and reusable without leaving natural language.

\section{Methodology}
\label{Methodology}
This section presents the design and structure of PromptMN. It opens with the rationale for giving prompts a more defined form, then describes the syntax and keyword set that make up the language, explains how those keywords are grouped into functional clusters, and closes with the contract that governs how a model is expected to process PromptMN directives.

\subsection{Design Rationale}

Prompt engineering has been developing from informal experimentation into a recognised discipline with documented techniques, including role prompting, chain-of-thought reasoning, few-shot demonstrations, and explicit constraints~\cite{ekin2026prompt, liu2026comprehensive}.Yet the practice has remained largely unstructured: prompts take the form of free-form natural language instructions, with a notable lack of common vocabulary, consistent formatting conventions, or any mechanism for systematic adaptation across tasks and contexts~\cite{li2025survey}. As prompts grow to coordinate multi-agent workflows, their implicit structure becomes a bottleneck, which motivates the case for integrating explicit prompting frameworks rather than relying on unstructured free-text instructions~\cite{tang2026llm}.

Context engineering research has made the cost of this informality visible. As observed by Philipp Schmid \cite{savva2026context}, a substantial subset of agent failures can be classified as context failures rather than fundamental model failures, underscoring the need for structured prompt markup. Hua et al.~\cite{hua2025context} trace context engineering from its roots in sensor-driven, rule-based HCI systems, where designers acted as "intention translators", through to today's agentic pipelines, where LLMs are still fundamentally bottlenecked by the quality of input tokens rather than model capacity alone. Work on building production agentic systems~\cite{bandara2025practical} traces brittle pipelines and non-deterministic failures back to weak task decomposition and agents carrying too many responsibilities, while ReAct~\cite{yao2022react} showed that weaving reasoning traces alongside actions drops hallucination rates and keeps decision steps auditable. Yet none of these efforts appears to provide a formal, reusable vehicle for expressing that structure.

PromptMN addresses this gap in part. It introduces a pseudo-prompting domain-specific language (DSL) whose reserved keywords impose sufficient structure for prompts to be parsed, reviewed, and reused, while leaving all prose content deliberately unconstrained. The goal is not to replace natural language but to annotate it with typed, mnemonic directives.

\subsection{Language Design}

PromptMN keywords are written inline with natural language and prefixed with `\%`. They act as typed directives that an LLM or agentic runtime is expected to recognise and resolve. Statements are terminated by semicolons (\texttt{;}), and related statements can be grouped into block scopes with curly braces (\texttt{\{\ldots\}}). Newlines (\texttt{\textbackslash n}) mark the start of a new term or expression.

A central design decision is semantic resolution over source order: the interpreter reorders directives by role before execution, so authors may compose prompts in any reading order without affecting interpretation. By relying on the natural language understanding capabilities of the underlying model, this approach relaxes the authoring burden while supporting a more predictable and well-defined semantic interpretation.

The vocabulary is intentionally small, comprising forty-eight named tokens plus the numeric step pattern \texttt{\%<number>}, and every token is mnemonic (\texttt{\%plan}, \texttt{\%req}, \texttt{\%if}, \texttt{\%out}). Keeping the surface area small lowers the entry cost for writers, echoing prior guidance that prompts work best when they are concise and unambiguous, while wordy or vague phrasing tends to produce weaker outputs~\cite{choi2025survey, ekin2026prompt}. The delimiter pair \texttt{$\infty$ \ldots $\infty$} marks the boundary of a PromptMN block, making it easy to embed programs inside larger documents or system prompts. Such explicit declarations also help an interpreter reliably recognise PromptMN keywords within surrounding natural-language context, avoiding accidental capture of similar words in the user's prose.

\subsection{Cluster Architecture}

\begin{figure}[h]
\centering
\includegraphics[width=1.0\textwidth]{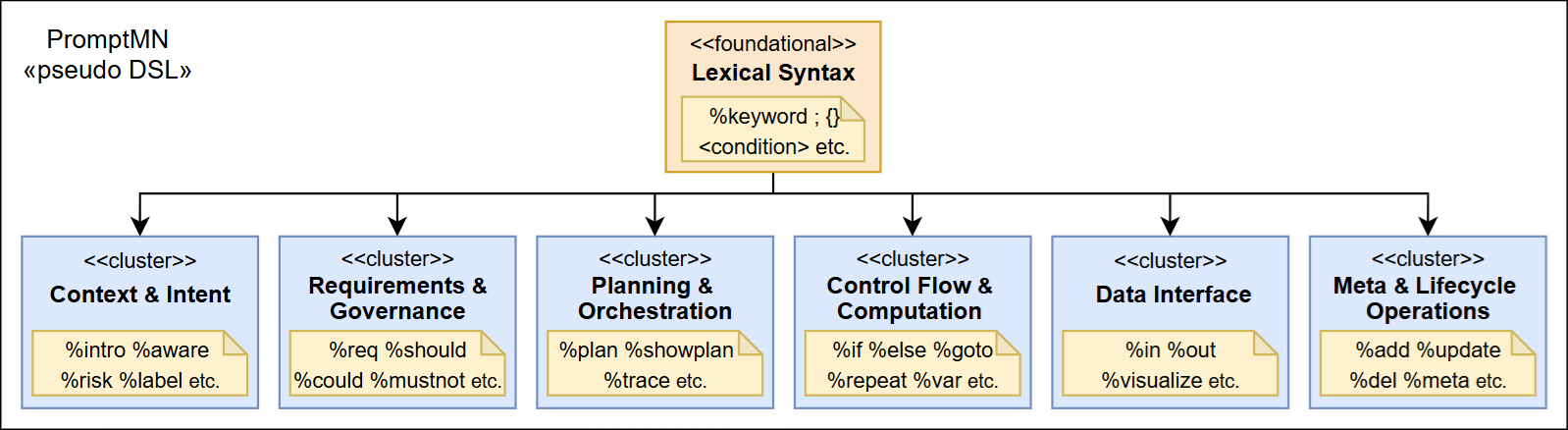}
\caption{Conceptual architecture of PromptMN. A pseudo‑prompting language (DSL) whose keywords are grouped into seven peer clusters: a foundational Lexical Syntax and six intent clusters covering Context \& Intent, Requirements \& Governance, Planning \& Orchestration, Control Flow \& Computation, Data Interface, and Meta \& Lifecycle Operations.}
\label{fig:conceptualarchitecture}
\end{figure}

The keywords are organised into seven peer clusters as shown in Figure~\ref{fig:conceptualarchitecture}. Cluster~1 constitutes the foundation, while Clusters~2--7 group directives by intent and broadly reflect the kinds of concerns that surface whenever a user collaborates with a GenAI model, whether for everyday business tasks or across the AI-assisted software development lifecycle.

\textbf{Cluster~1: Lexical Syntax} governs the entire prompt. It defines delimiters, block scope rules, keyword resolution order, and meta-syntactic placeholders. Every other cluster presupposes it.

\textbf{Cluster~2: Context \& Intent} (\texttt{\%role}, \texttt{\%intro}, \texttt{\%goal}, \texttt{\%aware}, \texttt{\%risk}, \texttt{\%problem}, \texttt{\%example}, \texttt{\%note}, \texttt{\%label},  \textit{\%domain)} captures the business and domain framing of a task. \texttt{\%role} hands the model one or more personas so the interpreter settles into the right stance before touching any other directive. \texttt{\%goal} states the intended success criterion, complementing \texttt{\%problem} (the gap) with a forward-looking target. \textit{\%domain} extends the context layer by scoping the model's frame of reference to a particular subject area, supplying the terminology and background knowledge that shape how every subsequent directive is interpreted. Together, these additions complete the intent layer: who is acting, toward what end, from what starting condition, and within which knowledge domain. Taken as a whole, the cluster handles the framing work that decides whether the output is even pointed in the right direction before any reasoning starts.

\textbf{Cluster~3: Requirements \& Governance} (\texttt{\%req}, \texttt{\%reqfunc}, \texttt{\%reqnonfunc}, \texttt{\%newreq}, \texttt{\%should}, \texttt{\%could}, \texttt{\%rule}, \texttt{\%mustnot}) encodes MoSCoW-style priority signals and hard constraints as typed keywords. Lifting constraints out of free prose and into first-class tokens makes them visible to tooling and harder for the model to quietly drop.

\textbf{Cluster~4: Planning \& Orchestration} (\texttt{\%plan}, \texttt{\%<number>}, \texttt{\%showplan}, \texttt{\%trace}) supports multi-step execution. \texttt{\%plan} states the high-level strategy; \texttt{\%<number>} then enumerates its concrete steps in ascending order (\texttt{\%1}, \texttt{\%2}, \texttt{\%3}, \ldots), letting authors scatter step definitions freely across the prompt while the interpreter reassembles them in sequence before acting. \texttt{\%showplan} forces that assembled plan to be surfaced before any action is taken, which keeps an early misstep from quietly cascading through later steps. \texttt{\%trace} asks for a post-hoc reasoning log delivered through \texttt{\%out}, giving reviewers something concrete to inspect after the fact.

\textbf{Cluster~5: Control Flow \& Computation} (\texttt{\%if}, \texttt{\%else}, \texttt{\%repeat}, \texttt{\%continue}, \texttt{\%break}, \texttt{\%goto}, \texttt{\%var}, \texttt{\%method}, \textit{\%return}) brings structured programming constructs into the prompt layer. Branching and iteration tend to creep into longer prompts, whether the author plans for them or not; turning them into typed constructs trims the ambiguity of English phrasing and opens the door to reuse through \texttt{\%var} substitution and \texttt{\%method} encapsulation. \textit{\%return} completes the method abstraction by allowing a \textit{\%method} to terminate at any point and pass a value back to its call site, enabling named behaviors to be composed into larger prompt programs.

\textbf{Cluster~6: Data Interface} (\texttt{\%in}, \textit{\%data} \textit{\%ignore},\texttt{\%out}, \texttt{\%visualize}, \texttt{\%diagram}) makes data boundaries explicit. Drawing a clean line between input and output is routine in software design; in PromptMN that line becomes a keyword-level contract rather than something the reader has to infer. \textit{\%data} extends this boundary by declaring named, schema-aware structured collections — records, tables, lists, or key-value sets — that can be referenced consistently across directives, methods, and steps. \textit{\%ignore} acts as the negative counterpart to \textit{\%in}, explicitly excluding specified files, directories, or context elements from the task scope so they are never read or factored into execution regardless of their presence in the context window.  \texttt{\%visualize} and \texttt{\%diagram} are output-type specifiers that constrain the response to visual artifacts.

\textbf{Cluster~7: Meta \& Lifecycle Operations} (\texttt{\%add}, \texttt{\%del}, \texttt{\%update}, \texttt{\%addition}, \texttt{\%newconcept}, \texttt{\%format}, \texttt{\%meta}) treats prompts as maintainable artefacts. \texttt{\%newconcept} lets the vocabulary stretch in-session; \texttt{\%meta} captures authorship, version, intended runtime, and locale. 

The original, full specification of PromptMN is demonstrated in Appendix~\ref{appendix:promptmn}, and its concise version, as distilled by Claude Opus 4.8, is presented in Appendix~\ref{appendix:promptmn-concise}.

\subsection{Interpreter Contract}

PromptMN does not prescribe a specific model or execution engine. The interpreter contract requires only that its specification be present within a context window or system prompt, or that an LLM be fine-tuned on PromptMN for its proper recognition. This loose coupling means any sufficiently capable LLM can serve as the runtime, and compliance can be tested with a small suite of functional prompts whose observable outputs, namely the externalized execution plan from \texttt{\%showplan} and the post-hoc reasoning log from \texttt{\%trace}, provide concrete, inspectable evidence of correct directive resolution.

The language is intentionally forward-compatible: \texttt{\%newconcept} lets practitioners introduce domain-specific keywords without modifying the base specification, and \texttt{\%meta} provides a hook for runtime-specific extensions. Both mechanisms reflect the observation that context engineering practices must evolve alongside model capabilities. Both mechanisms reflect the observation that Hua et al.~\cite{hua2025context} place at the center of their framework: the techniques and formats of context engineering must change as machine intelligence advances.

\section{Evaluation and Case Studies}
\label{CaseStudies}
The section evaluates PromptMN through practical scenarios, moving from basic constructs to realistic AI-assisted tasks. It first demonstrates how current AI models recognize and execute core PromptMN directives, then extends the discussion to reverse prompt engineering, where PromptMN offers a structured and effective approach for human-AI collaboration in the AI era.

\subsection{Structured Execution: From Basic Constructs to Development Workflows}

The following examples and case studies demonstrate PromptMN’s structural recognition by frontier AI models and its suitability for integrated task execution. Each prompt included the full PromptMN specification (Appendix~\ref{appendix:promptmn}) as context. Table~\ref{tab:ex_repeat} presents the simplest case: a \%repeat loop that outputs a fixed string three times.

\begin{table}[ht]
\caption{Basic loop: \texttt{\%repeat} with inline \texttt{\%out}.}
\centering
\begin{tabular}{p{0.45\textwidth} p{0.45\textwidth}}
\toprule
\textbf{Code} & \textbf{Output} \\
\midrule
\begin{minipage}[t]{0.45\textwidth}
\begin{verbatim}
Using PromptMN, complete the
following task:
  %repeat <3 times>
    %out: Hello AI World!
\end{verbatim}
\end{minipage}
&
\begin{minipage}[t]{0.45\textwidth}
\begin{verbatim}
Hello AI World!
Hello AI World!
Hello AI World!
\end{verbatim}
\end{minipage}
\\
\bottomrule
\end{tabular}
\label{tab:ex_repeat}
\end{table}

To verify that more complex control flow is handled reliably, Table~\ref{tab:ex_method} evaluates a prime-checking method across four leading models, each returning the correct result for the given input.

\begin{table}[h!]
\caption{Method definition: \texttt{\%method}, \texttt{\%var}, \texttt{\%if}, \texttt{\%repeat}}
\centering
\begin{tabular}{p{0.6\textwidth} p{0.3\textwidth}}
\toprule
\textbf{Code} & \textbf{Output} \\
\midrule
\begin{minipage}[t]{0.6\textwidth}
\begin{verbatim}
%method %is-prime(%n) {
    %if <%n less than 2>
        { %return false; }
    %var %counter = 2;
    %repeat <%counter is at most half of %n> {
        %if <%counter divides %n evenly>
            { %return false; }
        %counter = %counter + 1;
    }
    %return true;
}
%var %number = 23;
%out %is-prime(%number);
\end{verbatim}
\end{minipage}
&
\begin{minipage}[t]{0.3\textwidth}
\begin{verbatim}
Claude Fable 5:
true

Claude Opus 4.8:
true

Gemini 3.1 Pro:
true

GPT-5.5:
true
\end{verbatim}
\end{minipage}
\\
\bottomrule
\end{tabular}
\label{tab:ex_method}
\end{table}

\begin{figure}[ht]
\centering
\includegraphics[width=0.95\textwidth]{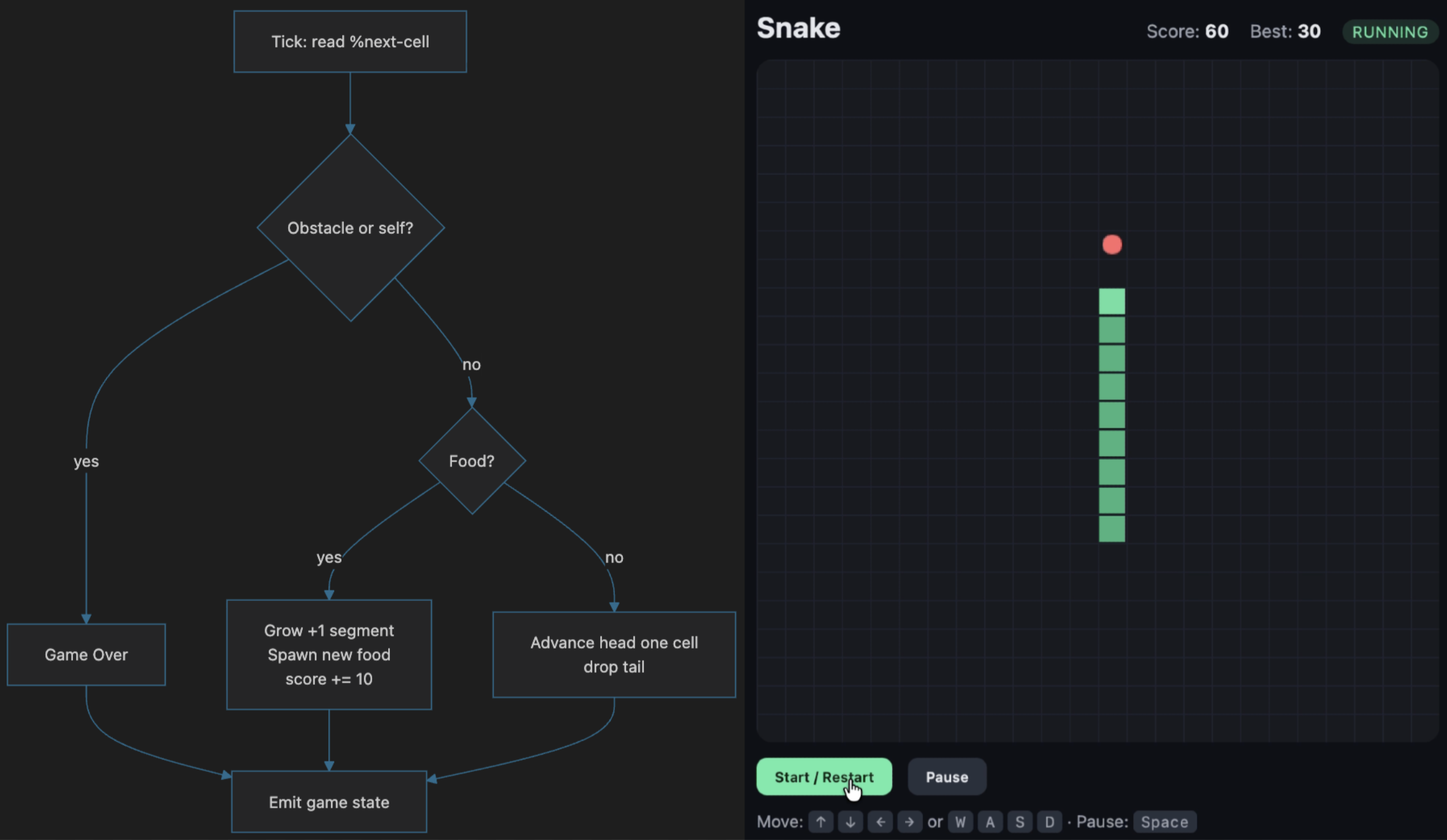}
\caption{Per-tick decision flow for the Snake game (left) and the result of the generated code running in a browser (right).}
\label{fig:if-else-snake-game}
\end{figure}

The example in Table\ref{tab:ex_ifelse} presents a short and complete instruction of a typical Snake game in PromptMN format for AI-assisted development or natural-language-driven programming ('vibe coding'), and explicitly shows the model's execution plan. Furthermore, Figure\ref{fig:if-else-snake-game} shows the model's decision flow alongside its running implementation in a browser.

\begin{table}[H]
\caption{Conditional branching: \texttt{\%if} and \texttt{\%else} chain that resolves one tick of a Snake game.}
\centering
\begin{tabular}{p{0.55\textwidth} p{0.35\textwidth}}
\toprule
\textbf{Code} & \textbf{Output} \\
\midrule
\begin{minipage}[t]{0.55\textwidth}
\begin{verbatim}
Using PromptMN, complete the following task:
%role game developer;
%intro a platform-independent Snake game; on
       every tick the snake advances by one cell
       in its current direction.
%goal decide what happens next based on the
      next cell the snake is about to enter.

%var %next-cell;
%if <%next-cell is an obstacle OR 
    is part of the snake’s body> {
    end the game and show "Game Over";
}
%else %if <%next-cell contains food> {
    grow the snake by one segment;
    spawn new food in a random empty cell;
    increment the score by 10;
}
%else {
    move the snake forward by one cell;
}
%showplan;
%out the resulting game state;
\end{verbatim}
\end{minipage}
&
\begin{minipage}[t]{0.35\textwidth}
\begin{verbatim}
Claude Opus 4.7:

Plan (`%showplan`)
As a `game developer`, I will 
resolve the prompt in this order:

1. Bind `%next-cell` — read the 
   cell the snake's head will 
   occupy on this tick.
2. Evaluate the branch conditions 
   in priority order:
   - Collision — `%next-cell` is 
     an obstacle OR overlaps 
     the snake's body.
   - Food — `%next-cell` contains food.
   - Empty — default movement.
3. Apply the matching branch's state 
   mutations atomically (single tick 
   = single transition).
4. Emit the resulting game state 
    per `%out`. ...
\end{verbatim}
(Figure~\ref{fig:if-else-snake-game}'s flow was also part of the prompt.)
\end{minipage}
\\
\bottomrule
\end{tabular}
\label{tab:ex_ifelse}
\end{table}

Table~\ref{tab:case_codebase} illustrates four representative scenarios that span the typical development lifecycle: creating a new codebase, maintaining an existing one, redesigning with architectural improvements, capturing requirements as a structured prompt, and mixing PromptMN directives with natural language.

\begin{table}[H]
\caption{Five representative codebase scenarios expressible in PromptMN.}
\centering
\begin{tabular}{p{0.16\textwidth} p{0.78\textwidth}}
\toprule
\textbf{Scenario} & \textbf{PromptMN sketch} \\
\midrule

(a) Create a new codebase &
\begin{minipage}[t]{0.78\textwidth}
\begin{verbatim}
%role full-stack engineer;
%goal scaffold a new Next.js project with authentication;
%req { TypeScript; Tailwind CSS; NextAuth.js; }
%plan initialise; wire auth; add CI;
%1 run create-next-app with TypeScript template;
%2 configure NextAuth with the GitHub provider;
%3 add a GitHub Actions workflow for lint + test;
%showplan;
%out file tree and the key configuration snippets;
\end{verbatim}
\end{minipage}
\\
\midrule

(b) Update an existing codebase &
\begin{minipage}[t]{0.78\textwidth}
\begin{verbatim}
%role maintainer;
%intro project: order-service, Java 21, Spring Boot;
%goal add idempotent retry to the payments endpoint;
%mustnot { change the existing public API; }
%req emit a structured log on every retry attempt;
%trace;
%out a unified diff against the main branch;
\end{verbatim}
\end{minipage}
\\
\midrule

(c) Restart with improvements &
\begin{minipage}[t]{0.78\textwidth}
\begin{verbatim}
%role solution architect;
%intro current app is a monolith with slow cold start;
%goal redesign as event-driven micro-services;
%req { sub-second cold start; OpenTelemetry tracing; }
%newreq replace internal REST calls with gRPC;
%plan greenfield design; migration path; deprecation;
%showplan;
%out a complete PromptMN instruction another
     engineer can hand back to an LLM and execute;
\end{verbatim}
\end{minipage}
\\
\midrule

(d) SRS expressed as a prompt &
\begin{minipage}[t]{0.78\textwidth}
\begin{verbatim}
%role business/system analyst;
%intro {
  Learning Management System (LMS): a web platform that creates,
  delivers, and tracks educational content and assessments for
  learners and instructors;
  Quiz Module: the LMS sub-system responsible for authoring,
  scheduling, delivering, and auto-grading online quizzes;
}
%goal capture an SRS for the quiz module of an LMS;
%reqfunc { create quiz; submit quiz; auto-grade; }
%reqnonfunc { 99.5% monthly uptime;
              p95 response time < 300 ms;
              GDPR compliant; }
%should partial-credit grading;
%could camera-based proctoring;
%rule { a quiz is visible only between its
        start and end timestamps; }
%format markdown;
%out SRS document with sections F-Req, NF-Req, Business Rules;

\end{verbatim}
\end{minipage}
\\
\midrule
(e) Mixed usage with PromptMN
 &
\begin{minipage}[t]{0.78\textwidth}
\begin{verbatim}
Please review this pull request and summarise the changes in plain 
English. %aware the codebase follows PEP-8 and avoids global state;  
%mustnot suggest renaming any public API; %showplan

\end{verbatim}
\end{minipage}
\\
\bottomrule
\end{tabular}
\label{tab:case_codebase}
\end{table}

\subsection{Reverse Prompt Engineering with PromptMN}

Reverse prompt engineering (RPE), also known as reverse prompting, flips the conventional workflow on its head: instead of crafting a prompt and hoping the model interprets it correctly, you supply the desired output and ask the model to work out what prompt would most reliably reproduce it. Rather than relying on a single, static instruction written upfront, this retrospective loop lets the model continuously refine its contextual understanding by reasoning backward from results. Femepid et al. demonstrate that this self-correcting dynamic leads to measurable, compounding improvements in output accuracy over time \cite{femepid2024gradual}.

RPE pairs particularly well with PromptMN. Because PromptMN prompts are structured, typed, and semantically ordered, with each keyword carrying an unambiguous intent, the model has a far cleaner signal to reason from than it would with free-form prose. Tracing which directive produced which part of the output becomes straightforward, making the backward-reasoning loop faster, more precise, and more actionable.

Representative examples of RPE are shown in Table ~\ref{tab:reverse-prompting}. The complete sets of completions generated by Claude Fable 5 for each prompt are provided in Appendices ~\ref{app:promptmn-penguin}, ~\ref{app:penguin-prompt-b}, and ~\ref{app:penguin-prompt-srs}, respectively. A comparison of the completions across the different reverse prompts reveals distinct characteristics. The completion of the simple prompt (b) reflects the LLM's default stylistic preferences; in other words, the model's predictive mechanism is directly reflected in the output. The SRS-styled reverse prompt (c) follows the given instruction and is generated in accordance with the SRS format, which was originally designed for human-to-human communication. In contrast, the PromptMN-based reverse prompt (a) demonstrates a structure explicitly designed for human-to-AI interaction.

\begin{table*}[h]
\caption{Reverse Prompting Examples}
\centering
\begin{tabular}{p{0.16\textwidth} p{0.78\textwidth}}
\toprule
\textbf{Scenario} & \textbf{User Request} \\
\midrule

(a) PromptMN-based prompt generation &
\begin{minipage}[t]{0.78\textwidth}
\begin{verbatim}
I would like to create a super-fun, platform-independent 2D retro action 
game with zero platform dependencies. The core elements will include a 
penguin-styled player character, alongside basic enemy types such as 
walkers, flyers, shooters, and chargers. Based on the PromptMN language, 
could you write a comprehensive prompt that covers all typical game 
features for a third party who will use your refined prompt for AI-assisted 
coding?
\end{verbatim}
\end{minipage}
\\
\midrule

(b) Simple Prompt generation &
\begin{minipage}[t]{0.78\textwidth}
\begin{verbatim}
I would like to create a super-fun, platform-independent 2D retro action
game with zero platform dependencies. The core elements will include 
a penguin-styled player character, alongside basic enemy types such as 
walkers, flyers, shooters, and chargers. Could you write a comprehensive 
prompt that covers all typical game features, which a third party can then 
use for AI-assisted coding?
\end{verbatim}
\end{minipage}
\\
\midrule

(c) SRS generation &
\begin{minipage}[t]{0.78\textwidth}
\begin{verbatim}
I would like to create a super-fun, platform-independent 2D retro action 
game with zero platform dependencies. The core elements will include a 
penguin-styled player character, alongside basic enemy types such as 
walkers, flyers, shooters, and chargers. Could you write a comprehensive 
Software Requirements Specification covering all features for a third party 
who will use your refined prompt for AI-assisted coding?
\end{verbatim}
\end{minipage}
\\
\bottomrule
\end{tabular}
\label{tab:reverse-prompting}
\end{table*}

Together, these examples indicate that PromptMN supports AI interaction across multiple levels, from executable directives to broader development and requirements-oriented tasks. Its structured form improves interpretability while preserving the flexibility of natural language. These findings suggest that PromptMN can serve as a practical notation for clearer, more reliable human-AI communication, as discussed in the concluding section.

\section{Conclusion}
\label{Conclusion}
PromptMN proposes a compact domain-specific language for a problem that now sits at the centre of human-to-AI interaction: how to express intent clearly enough that a generative model can act on it, preserve it, and expose how it was interpreted. Prompting is no longer only a matter of writing persuasive natural-language instructions. In agentic and software-development settings, a prompt increasingly behaves like an executable handoff. It carries role, goal, context, constraints, priorities, data boundaries, planning steps, and expected outputs. When these elements remain buried in free-form prose, an early misunderstanding can move through the workflow before a human can see where the drift began.

The contribution of PromptMN is to make that handoff explicit without asking users to become programmers. Its \texttt{\%}-prefixed directives provide a small shared vocabulary for structuring ordinary language: \texttt{\%role} establishes stance, \texttt{\%goal} defines the target, \texttt{\%req}, \texttt{\%should}, \texttt{\%could}, and \texttt{\%mustnot} make priorities visible, \texttt{\%plan} and \texttt{\%showplan} surface the intended sequence of work, and \texttt{\%in} and \texttt{\%out} clarify data boundaries and expected results. The language does not replace prose. It gives prose a typed frame, so both the author and the model can inspect the same instruction with less ambiguity.

\subsection*{Impact on Human-to-AI Interaction}

Typed, semantically ordered directives give a model a cleaner signal than loose prose. They make it easier to see what the user meant and which part of the instruction shaped which part of the response. \texttt{\%showplan} externalizes the assembled plan before work begins, while \texttt{\%trace} can return a reasoning log afterward. Together, they create review points at both ends of the exchange. Because the vocabulary is small and reads as plain annotation, PromptMN is intended to make structured intent more approachable beyond specialist users, including analysts, managers, developers, and everyday users of generative tools.

\subsection*{Practical Implications for the SDLC}

PromptMN fits software development especially well because SDLC professionals already think in structured units such as requirements, constraints, plans, interfaces, and outputs. At the same time, the notation lowers the entry cost for less technical stakeholders because it works like a lightweight template for what a good prompt should contain. In the scenarios examined here, the same vocabulary described a fresh codebase, an update to an existing system, an architectural redesign, an SRS-style prompt, and a mixed natural-language request with only a few directives attached. A full PromptMN block can guide a complex task, while single directives such as \texttt{\%aware}, \texttt{\%mustnot}, or \texttt{\%showplan} can strengthen an ordinary request when a full block would be unnecessary.

The demonstrated results indicate that PromptMN is already recognizable to current advanced models. Empirical testing indicates that contemporary models, such as Claude Fable 5, Claude Opus 4.8, Gemini 3.1 Pro, and GPT-5.5, reliably parse and execute PromptMN directives out of the box, and this confirms that the syntax is viable without the overhead of specialized fine-tuning. The Snake game example further shows how a surfaced plan can be reviewed before code or state changes are produced. For AI-assisted development, this matters because early visibility can reduce needless iteration and prevent an incorrect interpretation from becoming rework later in the chain.

PromptMN also pairs naturally with reverse prompt engineering. When a model restates a desired outcome as a PromptMN prompt, the result becomes a structured artifact that separates role, goal, requirements, constraints, format, and output expectations. Comparing a user's original prompt with the model's PromptMN-style reverse prompt can reveal missing assumptions, weak framing, or constraints the model failed to infer. This makes reverse prompting more useful when the prompt will be reused by another person, another model, or an agentic workflow.

\subsection*{Limitations and Future Work}

The results in this paper should be read as demonstrative rather than conclusive. PromptMN has been shown to work in selected examples and SDLC-oriented scenarios, but it has not yet been validated through a controlled benchmark, a user study, or a large-scale comparison against plain prompting, SRS-style prompting, and programming-style pseudocode prompting. Its current interpreter is also probabilistic. Unless a model is fine-tuned or a runtime enforces the specification, correct interpretation depends on the model, the context window, and the prompt environment.

Prompt injection is also worth noting: if PromptMN directives appear in a system prompt and user input is concatenated into the same context, a user could insert keywords to alter the intended behavior. Input sanitization and clear separation between system and user context are practical mitigations that future work should evaluate.

There is also a practical cost. If the full PromptMN specification in Appendix ~\ref{appendix:promptmn} must be attached to every prompt, it consumes additional tokens. Possible responses include fine-tuning a model on the specification, using a concise model-facing version in Appendix ~\ref{appendix:promptmn-concise}, or building a linker that attaches only the definitions of the keywords used in a particular prompt. These approaches could reduce token overhead, but they depend on model capability and should be evaluated carefully rather than assumed to be equivalent to the full specification.

Future work should therefore focus on formalization, measurement, and tooling. PromptMN needs a grammar, parser, and linter so that structural errors can be detected before a prompt reaches the model. Its effectiveness should be measured across models, domains, prompt lengths, and user groups, using metrics such as task success, constraint following, output variance, time saved, and user effort. Since PromptMN is introduced here as a single-author proposal, its further development should also remain open to contributions from researchers, practitioners, and users who can test, adapt, and improve it in real workflows. Practical tooling should bring PromptMN into everyday use through syntax highlighting, prompt validation, prompt diffing, reverse-prompt support, and tested token-saving strategies such as fine-tuned interpreters, concise specifications, and keyword linkers.

Overall, PromptMN is best understood as a communication layer between free-form natural language and formal programming. Its promise is not that it removes uncertainty from generative AI, but that it makes intent, structure, and review points more visible. For human-to-AI interaction in general, and for SDLC professionals in particular, that visibility turns prompting from a fragile paragraph into a compact, inspectable, and reusable instruction artifact.

\section{Acknowledgments}
I sincerely thank my former and current colleagues around the world for their support and thoughtful discussions. The AI models referenced in this paper have also been helpful in various ways, including proofreading. I am grateful to their creators, developers, and facilitators.

\bibliographystyle{unsrt}  
\bibliography{references}  

\appendix
\clearpage
\section{Appendix. The specification of the PromptMN}
\label{appendix:promptmn}
PromptMN is a pseudo\mbox{-}prompting language (domain\mbox{-}specific language) designed to provide typed, semantic directives for AI prompt interpretation and execution.

Version: 1.0.0,
Author: Enkhzol Dovdon,
License: MIT License,
Copyright (c) 2026 Enkhzol Dovdon

PromptMN defines a set of reserved keywords that function as interpreter directives. These directives govern how an AI model interprets, organises, and executes prompt instructions.

PromptMN annotates prompts using keywords written inline with prose, prefixed with \%, and resolved semantically before execution. This allows authors to compose prompts in any reading order without affecting interpretation.

The delimiter pair \(\infty\) \ldots{} \(\infty\) marks the program boundary and encloses the language\textquoteright{}s complete token set. Within this boundary, tokens are separated by commas (,), and every listed token carries a distinct meaning that the interpreter must correctly recognise and apply.

Keywords without an explicit format use the default forms \%keyword \textless{}text\textgreater{}; or \%keyword \{statements;\}.

\textbf{Token Set:}

\begin{quote}
\small\ttfamily\noindent
\(\infty\);,\allowbreak\{,\allowbreak\},\allowbreak\textbackslash{}n,\allowbreak\%role,\allowbreak\%intro,\allowbreak\%goal,\allowbreak\%techs,\allowbreak\%aware,\allowbreak\%risk,\allowbreak\%problem,\allowbreak\%example,\allowbreak\%note,\allowbreak\%label,\allowbreak\%domain,\allowbreak\%req,\allowbreak\%reqfunc,\allowbreak\%reqnonfunc,\allowbreak\%newreq,\allowbreak\%should,\allowbreak\%could,\allowbreak\%optional,\allowbreak\%rule,\allowbreak\%mustnot,\allowbreak\%plan,\allowbreak\%number,\allowbreak\%showplan,\allowbreak\%trace,\allowbreak\%if,\allowbreak\%else,\allowbreak\%repeat,\allowbreak\%continue,\allowbreak\%break,\allowbreak\%goto,\allowbreak\%var,\allowbreak\%method,\allowbreak\%return,\allowbreak\%in,\allowbreak\%data,\allowbreak\%ignore,\allowbreak\%out,\allowbreak\%visualize,\allowbreak\%diagram,\allowbreak\%add,\allowbreak\%del,\allowbreak\%update,\allowbreak\%addition,\allowbreak\%newconcept,\allowbreak\%format,\allowbreak\%meta\(\infty\)
\end{quote}

The keywords are organised into seven contextual clusters. Cluster 1 applies to the entire prompt as its foundation.

\subsection*{Cluster 1. Lexical Syntax (Foundational)}

\%keyword is generic notation for a unique command that provides additional instructional information. Keywords may appear multiple times and in any position within a prompt. Therefore, the interpreter should logically reorder keywords, resolving them by semantics rather than source order, before executing or following their instructions.

Where the syntax is non-obvious, a Format and a For example line follow the definition; otherwise, the default keyword forms given above apply.

Semicolon (;) completes an expression statement.

Curly braces (\{ ... \}) define a block scope by grouping a sequence of statements that belong to the command that immediately precedes them; they are written and used as the pair \{ ... \}, and the entries \{ and \} in the enumeration above denote the opening and closing of that pair. Format: \%keyword \{statement; statement; ...\}

\textbackslash{}n (newline) indicates a line break. After \textbackslash{}n, a new term or expression begins.

\textless{}condition\textgreater{} represents a condition that can be evaluated as true or false, or as satisfied or not satisfied. A condition can be a simple declaration or a logical expression.

Angle-bracket tokens such as \textless{}condition\textgreater{}, \textless{}name\textgreater{}, or \textless{}expression\textgreater{} are meta-syntactic placeholders and are not themselves registered keywords.

\subsection*{Cluster 2. Context and Intent}

\%role assigns one or more personas to the AI model or agent, instructing it to interpret and respond as the specified role(s). All subsequent directives in the prompt are processed through this persona lens. Format: \%role \textless{}role\textgreater{}; or \%role \{role1; role2; ...\} For example: \%role senior software architect; or \%role \{senior software engineer; AI engineer;\}

\%intro background information or contextual overview necessary to understand the problem, domain, or document scope.

\%goal defines the intended purpose or success criteria of the task and explains what a good outcome looks like, distinct from \%problem, which describes the current unsatisfactory situation. Format: \%goal \textless{}text\textgreater{}; or \%goal \{statements;\}

\%techs declares the preferred technology stack(s) for the solution; any recommendation or design decision must favour these technologies unless a constraint prevents it.

\%aware indicates information, assumptions, risks, or conditions that the reader or system should be aware of when interpreting or applying the requirements.

\%risk flags a potential event that may negatively affect the outcome and should therefore be accounted for in the proposal or decision.

\%problem states the situation or need to be addressed, expressed as what is wrong or missing before deciding on a solution.

\%example provides a supporting illustration (sample scenario or input/output) used to clarify meaning but not to replace real requirements.

\%note records a brief supporting point used to clarify or highlight information.

\%label assigns a short name or tag used to identify, categorize, or describe an item quickly.

\%domain specifies the applicable domain and provides relevant knowledge, terminology, and background context from that field that the AI model or agent should draw upon when interpreting and executing the prompt. It scopes the model's frame of reference to a particular subject area, ensuring that domain-specific conventions and assumptions are applied consistently across all subsequent directives.

\subsection*{Cluster 3. Requirements \& Governance}

\%req states a testable requirement of what it must do or satisfy.

\%reqfunc describes a functional requirement: a behavior or function the system must perform (what it does).

\%reqnonfunc specifies a non-functional requirement: a quality attribute or constraint, such as performance, security, or reliability (how well it works).

\%newreq introduces a new requirement, and if there is a conflict between the new one and the current requirement(s) in the existing context, the new requirement overrides them. Unlike \%update, which performs a generic edit on any element, \%newreq is a typed override scoped to requirements.

\%should marks a Should-have: important but not vital.

\%could marks a Could-have: nice to have; deliver it only if time and resources remain.

\%optional marks item(s) the solution should incorporate unless doing so conflicts with or disrupts another requirement or constraint.

\%rule defines a condition-based policy or logic that determines what is allowed, required, or triggered in specific situations.

\%mustnot imposes a hard constraint that restricts options or forbids certain behaviors, technologies, or conditions. Format: \%mustnot \{text1; text2; \ldots{}\} For example: \%mustnot do not hallucinate

\subsection*{Cluster 4. Planning \& Orchestration}

\%plan presents instructions or communicates one or more intentions to the AI model or agent. It states the intended outcome or high-level strategy rather than the program itself.

\%\textless{}number\textgreater{} declares a sequentially ordered step or event, where the integer indicates execution position. Steps are processed in ascending numeric order and may appear anywhere in the prompt; the interpreter collects and sequences them before execution. In the token set, \%number denotes this abstract category; concrete steps are written as \%1, \%2, \%3, and so on. Format: \%1 \textless{}text\textgreater{}; \%2 \textless{}text\textgreater{}; \%3 \textless{}text\textgreater{}; For example: \%1 load dataset; \%2 preprocess features; \%3 train model; \%4 evaluate metrics;

\%showplan specifies that the AI model or agent must explicitly present its execution plan before performing any actions.

\%trace requires the AI model or agent to emit an execution trace or reasoning summary alongside its output, so that the steps actually taken can be reviewed after the fact. The trace itself is delivered via \%out and complements \%showplan, which is presented before the actions.

\subsection*{Cluster 5. Control-Flow \& Computation}

\%if \textless{}condition\textgreater{} if the condition evaluates to true or valid, a statement or a group of statements defined within curly braces should be followed or executed accordingly. If the condition is false or invalid, skip the statements (or perform an alternative action). Format: \%if \textless{}condition\textgreater{} \{statements;\} or \%if \textless{}condition\textgreater{} statement; For example: \%if \textless{}login is successful\textgreater{} \{navigate to the Home page;\}

\%else Use else to define an alternative action for the preceding if statement.
Format: \%else \{statements;\} or \%else statement; For example: \%if \textless{}login is successful\textgreater{} \{navigate to the Home page;\} \%else \{show the error message and offer sign-up;\}

\%repeat if the condition evaluates to true or valid, a statement or a group of statements defined within curly braces should be executed repeatedly until the condition becomes false or invalid. When the condition is false, stop the repetition. Format: \%repeat \textless{}condition\textgreater{} \{statements;\} or \%repeat \textless{}condition\textgreater{} statement; For example: \%repeat \textless{}there is no obstacle\textgreater{} the player should continue its current movement;

\%continue skips the remaining statements in the current iteration of the enclosing \%repeat loop and proceeds to re-evaluate its condition for the next iteration. Format: \%continue; For example: \%repeat \textless{}there are items\textgreater{} \{\%if \textless{}item is invalid\textgreater{} \{\%continue;\} process the item;\}

\%break terminates the enclosing \%repeat loop immediately, regardless of whether its condition still holds, and resumes execution after the loop. Format: \%break; For example: \%repeat \textless{}there are items\textgreater{} \{\%if \textless{}target is found\textgreater{} \{\%break;\} inspect the next item;\}

\%goto indicates that further processing should continue at the specified labeled statement, such as \%jumplabel-1, \%jumplabel-2, or \%jumplabel-3. A label is declared in-line using the form \%jumplabel-\textless{}N\textgreater{}: before the target statement, distinct from \%label in Cluster 2 which tags content. Format: \%jumplabel-\textless{}N\textgreater{}: statement; ... \%goto \%jumplabel-\textless{}N\textgreater{}; For example: \%jumplabel-1: retry the request; \%if \textless{}failed\textgreater{} \{\%goto \%jumplabel-1;\}

\%var declares a named data holder whose value can change while an algorithm or system operates. It assigns a value, and can also serve as a placeholder that is substituted with its value wherever it appears in the surrounding text, so the same prompt skeleton can be reused, localized, or refactored without manual editing. Within a string, any \%\textless{}name\textgreater{} token is substituted with its current value. Format: \%var \%\textless{}name\textgreater{}; \%\textless{}name\textgreater{} = \textless{}expression\textgreater{}; For example: \%var \%counter; \%counter = 1000; or \%var \%customer = "Alice"; \%out "Dear \%customer, your order is confirmed.";

\%method defines a named unit of behavior. Format: \%method \%\textless{}name\textgreater{}(\textless{}params\textgreater{}) \{statements;\} For example: \%method \%greet(\%user) \{\%out "Hello, \%user";\}

\%return terminates the current \%method execution and optionally passes a value or the result of an expression back to the call site. When used without a value, it exits the method immediately and returns control to the caller with no result. Format: \%return; or \%return \textless{}expression\textgreater{}; For example: \%method \%max(\%a, \%b) \{\%if \textless{}\%a \textgreater{} \%b\textgreater{} \{\%return \%a;\} \%return \%b;\}

\subsection*{Cluster 6. Data Interface}

\%in provides input data (files, structured data, or unstructured data) for an AI model or process.

\%data declares a named, structured collection --- such as a record, table, list, or key-value set that the AI model or agent should treat as a typed, addressable data structure rather than free-form text. It can be referenced by \%in, \%out, \%method, or step directives, enabling structured data to be passed and consumed consistently across the prompt program. Format: \%data \%\textless{}name\textgreater{} \{key: value; key: value; ...\} or \%data \%\textless{}name\textgreater{} [item1; item2; ...] For example: \%data \%user \{name: "Frank"; role: "admin"; active: true;\} or \%data \%scores [90; 85; 92;]

\%ignore instructs the AI model or agent to exclude and disregard the specified file(s), directories (including all their contents), or context elements from the task. Items listed under \%ignore are never read, referenced, or factored into execution, regardless of whether they appear elsewhere in the context window. Format: \%ignore \textless{}path\textgreater{}; or \%ignore \{path1; path2; ...\} For example: \%ignore \{draft/; example.md;\}

\%out declares the observable output.

\%visualize instructs the AI model or agent to render the specified subject as a visual artifact rather than (or in addition to) prose. \%visualize and \%diagram are specializations of \%out that constrain the output medium. Format: \%visualize \textless{}subject\textgreater{} \{format: \textless{}name\textgreater{}; statements;\} or \%visualize \textless{}subject\textgreater{}; For example: \%visualize \textless{}user login flow\textgreater{} \{format: mermaid; type: sequenceDiagram;\}

\%diagram is a shorthand for \%visualize using Mermaid as the default format. Format: \%diagram \textless{}subject\textgreater{} \{type: \textless{}mermaid-type\textgreater{};\} or \%diagram \textless{}subject\textgreater{}; For example: \%diagram \textless{}component architecture\textgreater{} \{type: flowchart;\}

\subsection*{Cluster 7. Meta \& Lifecycle Operations}

\%add introduces a new element into the current context.

\%del deletes an existing element from the current context.

\%update modifies an existing element in the current context.

\%addition used to provide additional information, with the context written using prompt syntax.

\%newconcept introduces a new keyword together with its explanation into the current context. If the new keyword conflicts with an existing one, the new definition overrides the existing one. Distinct from \%add, which introduces a content element; \%newconcept introduces a language element (a new keyword). Format: \%newconcept \%\textless{}keyword\textgreater{} \{explanation;\}

\%format declares a default rendering format for \%out or \%visualize within its scope, such as markdown, json, yaml, or plain text. Format: \%format \textless{}name\textgreater{}; For example: \%format markdown;

\%meta declares metadata about the PromptMN program itself rather than about the domain it describes, such as author, version, intended execution target, audience, or locale. \%meta is descriptive by default; agents may use it to gate or adapt execution (for example, refusing to run if the declared target does not match), but it has no built-in side effects. Format: \%meta \{key: value; key: value; ...\} For example: 

\clearpage
\section{Appendix. The specification of the PromptMN (Concise)}
\label{appendix:promptmn-concise}
PromptMN v1.0.0 --- a pseudo-prompting DSL giving typed, semantic directives for AI prompt interpretation/execution.

Author: Enkhzol Dovdon,
License: MIT License,
Copyright (c) 2026 Enkhzol Dovdon

\subsection*{Syntax}

\begin{itemize}
  \item Keywords are inline directives prefixed with \%, resolved semantically (not by source order). Authors may write keywords in any order/position; interpreter reorders by meaning before executing.
  \item Default forms: \texttt{\%keyword \textless{}text\textgreater{};} or \texttt{\%keyword \{statement; statement; ...\}}. Special forms noted per keyword.
  \item Program boundary: \texttt{\(\infty\) ... \(\infty\)} encloses the full token set; tokens comma-separated; each token has a distinct meaning the interpreter must apply.
\end{itemize}

\textbf{Token set:}

\begin{quote}
\small\ttfamily\noindent
\(\infty\);,\allowbreak\{,\allowbreak\},\allowbreak\textbackslash{}n,\allowbreak\%role,\allowbreak\%intro,\allowbreak\%goal,\allowbreak\%techs,\allowbreak\%aware,\allowbreak\%risk,\allowbreak\%problem,\allowbreak\%example,\allowbreak\%note,\allowbreak\%label,\allowbreak\%domain,\allowbreak\%req,\allowbreak\%reqfunc,\allowbreak\%reqnonfunc,\allowbreak\%newreq,\allowbreak\%should,\allowbreak\%could,\allowbreak\%optional,\allowbreak\%rule,\allowbreak\%mustnot,\allowbreak\%plan,\allowbreak\%number,\allowbreak\%showplan,\allowbreak\%trace,\allowbreak\%if,\allowbreak\%else,\allowbreak\%repeat,\allowbreak\%continue,\allowbreak\%break,\allowbreak\%goto,\allowbreak\%var,\allowbreak\%method,\allowbreak\%return,\allowbreak\%in,\allowbreak\%data,\allowbreak\%ignore,\allowbreak\%out,\allowbreak\%visualize,\allowbreak\%diagram,\allowbreak\%add,\allowbreak\%del,\allowbreak\%update,\allowbreak\%addition,\allowbreak\%newconcept,\allowbreak\%format,\allowbreak\%meta\(\infty\)
\end{quote}

Keywords are grouped in 7 clusters. Cluster 1 underlies the whole prompt.

\subsection*{Cluster 1 --- Lexical Syntax (foundational)}

\begin{itemize}
  \item \texttt{\%keyword}: generic notation for a unique command adding instructional info. May appear multiple times, any position; interpreter resolves by semantics, not order.
  \item \texttt{;} ends a statement.
  \item \texttt{\{ ... \}} block scope: groups statements belonging to the immediately preceding command. Tokens \texttt{\{} and \texttt{\}} are the open/close. Form: \texttt{\%keyword \{stmt; stmt; ...\}}.
  \item \texttt{\textbackslash{}n} newline: line break; a new term/expression begins after it.
  \item \texttt{\textless{}condition\textgreater{}}: something evaluable as true/false or satisfied/not (simple declaration or logical expression).
  \item Angle-bracket tokens (\texttt{\textless{}condition\textgreater{}}, \texttt{\textless{}name\textgreater{}}, \texttt{\textless{}expression\textgreater{}}) are meta-syntactic placeholders, NOT keywords.
\end{itemize}

\subsection*{Cluster 2 --- Context and Intent}

\begin{itemize}
  \item \texttt{\%role}: assigns persona(s); all later directives interpreted through this lens. \texttt{\%role \textless{}role\textgreater{};} or \texttt{\%role \{r1; r2;\}}. Ex: \texttt{\%role senior software architect;}
  \item \texttt{\%intro}: background/contextual overview of problem, domain, or scope.
  \item \texttt{\%goal}: intended purpose/success criteria; what good looks like (distinct from \%problem). \texttt{\%goal \textless{}text\textgreater{};} or \texttt{\%goal \{stmts;\}}.
  \item \texttt{\%techs}: preferred tech stack(s); recommendations must favor these unless a constraint prevents it.
  \item \texttt{\%aware}: info/assumptions/risks/conditions to be aware of when applying requirements.
  \item \texttt{\%risk}: a potential event that may harm the outcome; account for it.
  \item \texttt{\%problem}: the situation/need --- what is wrong or missing, before choosing a solution.
  \item \texttt{\%example}: supporting illustration (scenario or I/O) clarifying meaning; not a substitute for real requirements.
  \item \texttt{\%note}: brief supporting point to clarify/highlight.
  \item \texttt{\%label}: short name/tag to identify, categorize, or describe quickly.
  \item \texttt{\%domain}: specifies the domain and supplies its terminology/conventions/background; scopes the model's frame of reference for all later directives.
\end{itemize}

\subsection*{Cluster 3 --- Requirements \& Governance}

\begin{itemize}
  \item \texttt{\%req}: a testable requirement of what must be done/satisfied.
  \item \texttt{\%reqfunc}: functional requirement --- a behavior the system must perform.
  \item \texttt{\%reqnonfunc}: non-functional requirement --- quality/constraint (performance, security, reliability).
  \item \texttt{\%newreq}: introduces a new requirement; on conflict it overrides existing requirement(s). Typed override scoped to requirements (vs \%update's generic edit).
  \item \texttt{\%should}: Should-have --- important but not vital.
  \item \texttt{\%could}: Could-have --- nice to have; deliver only if time/resources remain.
  \item \texttt{\%optional}: item(s) to incorporate unless they conflict with another requirement/constraint.
  \item \texttt{\%rule}: condition-based policy defining what is allowed/required/triggered in given situations.
  \item \texttt{\%mustnot}: hard constraint forbidding behaviors/technologies/conditions. \texttt{\%mustnot \{t1; t2;\}}. Ex: \texttt{\%mustnot do not hallucinate}.
\end{itemize}

\subsection*{Cluster 4 --- Planning \& Orchestration}

\begin{itemize}
  \item \texttt{\%plan}: states intended outcome/high-level strategy (not the program itself).
  \item \texttt{\%\textless{}number\textgreater{}}: a sequentially ordered step; integer = execution position. Steps may appear anywhere; interpreter collects and sequences ascending before execution. \texttt{\%number} is the abstract category; concrete steps are \texttt{\%1, \%2, ...}. Ex: \texttt{\%1 load dataset; \%2 preprocess; \%3 train; \%4 evaluate;}
  \item \texttt{\%showplan}: model must present its execution plan before acting.
  \item \texttt{\%trace}: model must emit an execution trace / reasoning summary with its output (delivered via \%out); complements \%showplan (which comes before actions).
\end{itemize}

\subsection*{Cluster 5 --- Control-Flow \& Computation}

\begin{itemize}
  \item \texttt{\%if \textless{}condition\textgreater{}}: if true, execute statement/block; if false, skip (or do alternative). \texttt{\%if \textless{}cond\textgreater{} \{stmts;\}} or \texttt{\%if \textless{}cond\textgreater{} stmt;}. Ex: \texttt{\%if \textless{}login is successful\textgreater{} \{navigate to Home;\}}
  \item \texttt{\%else}: alternative for the preceding if. \texttt{\%else \{stmts;\}} or \texttt{\%else stmt;}.
  \item \texttt{\%repeat \textless{}condition\textgreater{}}: while true, execute statement/block repeatedly until false. \texttt{\%repeat \textless{}cond\textgreater{} \{stmts;\}} or \texttt{\%repeat \textless{}cond\textgreater{} stmt;}.
  \item \texttt{\%continue}: skip rest of current \%repeat iteration; re-evaluate condition. \texttt{\%continue;}
  \item \texttt{\%break}: terminate enclosing \%repeat immediately; resume after loop. \texttt{\%break;}
  \item \texttt{\%goto}: continue at a labeled statement. Labels declared inline as \texttt{\%jumplabel-\textless{}N\textgreater{}:} before the target (distinct from \%label). \texttt{\%jumplabel-\textless{}N\textgreater{}: stmt; ... \%goto \%jumplabel-\textless{}N\textgreater{};}. Ex: \texttt{\%jumplabel-1: retry; \%if \textless{}failed\textgreater{} \{\%goto \%jumplabel-1;\}}
  \item \texttt{\%var}: declares a mutable named value; also a placeholder substituted by its value wherever \texttt{\%\textless{}name\textgreater{}} appears in text. \texttt{\%var \%\textless{}name\textgreater{}; \%\textless{}name\textgreater{} = \textless{}expr\textgreater{};}. Ex: \texttt{\%var \%customer = "Alice"; \%out "Dear \%customer, your order is confirmed.";}
  \item \texttt{\%method}: defines a named unit of behavior. \texttt{\%method \%\textless{}name\textgreater{}(\textless{}params\textgreater{}) \{stmts;\}}. Ex: \texttt{\%method \%greet(\%user) \{\%out "Hello, \%user";\}}
  \item \texttt{\%return}: ends current \%method, optionally returning a value/expression; without a value, exits with no result. \texttt{\%return;} or \texttt{\%return \textless{}expr\textgreater{};}. Ex: \texttt{\%method \%max(\%a,\%b)\{\%if \textless{}\%a \textgreater{} \%b\textgreater{}\{\%return \%a;\} \%return \%b;\}}
\end{itemize}

\subsection*{Cluster 6 --- Data Interface}

\begin{itemize}
  \item \texttt{\%in}: provides input data (files, structured, or unstructured).
  \item \texttt{\%data}: declares a named structured collection (record/table/list/key-value) treated as typed addressable data, not free text; referenced by \%in/\%out/\%method/steps. \texttt{\%data \%\textless{}name\textgreater{} \{k: v; ...\}} or \texttt{\%data \%\textless{}name\textgreater{} [i1; i2; ...]}. Ex: \texttt{\%data \%user \{name:"Frank"; role:"admin"; active:true;\}}
  \item \texttt{\%ignore}: exclude file(s)/directories(+contents)/context elements; never read or factored in, even if present elsewhere. \texttt{\%ignore \textless{}path\textgreater{};} or \texttt{\%ignore \{p1; p2;\}}. Ex: \texttt{\%ignore \{draft/; example.md;\}}
  \item \texttt{\%out}: declares the observable output.
  \item \texttt{\%visualize}: render subject as a visual artifact (instead of/in addition to prose); specialization of \%out constraining the medium. \texttt{\%visualize \textless{}subject\textgreater{} \{format:\textless{}name\textgreater{}; stmts;\}} or \texttt{\%visualize \textless{}subject\textgreater{};}. Ex: \texttt{\%visualize \textless{}user login flow\textgreater{} \{format:mermaid; type:sequenceDiagram;\}}
  \item \texttt{\%diagram}: shorthand for \%visualize with Mermaid default. \texttt{\%diagram \textless{}subject\textgreater{} \{type:\textless{}mermaid-type\textgreater{};\}} or \texttt{\%diagram \textless{}subject\textgreater{};}.
\end{itemize}

\subsection*{Cluster 7 --- Meta \& Lifecycle Operations}

\begin{itemize}
  \item \texttt{\%add}: introduce a new element into context.
  \item \texttt{\%del}: delete an existing element from context.
  \item \texttt{\%update}: modify an existing element in context.
  \item \texttt{\%addition}: provide extra information, written in prompt syntax.
  \item \texttt{\%newconcept}: introduce a new keyword + explanation; overrides on conflict. Introduces a language element (vs \%add's content element). \texttt{\%newconcept \%\textless{}keyword\textgreater{} \{explanation;\}}
  \item \texttt{\%format}: default rendering format for \%out/\%visualize in scope (markdown, json, yaml, plain text). \texttt{\%format \textless{}name\textgreater{};}.
  \item \texttt{\%meta}: metadata about the PromptMN program itself (author, version, target, audience, locale). Descriptive by default; agents may use it to gate/adapt execution but it has no built-in side effects. \texttt{\%meta \{k: v; ...\}}
\end{itemize}

\clearpage
\section{Appendix. Reverse Prompt: PromptMN Program for the 2D Retro Penguin Action Game}
\label{app:promptmn-penguin}

\%meta \{author: game-design-lead; version: 1.0; target: any LLM coding agent; audience: AI-assisted developer; locale: en;\}

\%role \{senior game developer; 2D gameplay programmer; retro pixel-art designer;\}

\%domain 2D retro action / platformer games; apply established conventions: fixed-timestep game loop, sprite-based rendering, tile or bounding-box collision, entity state machines, arcade scoring, lives and power-ups, screen-shake and particle feedback ("game juice");

\%intro We are building a small but complete retro-style 2D action game from scratch. The protagonist is a penguin. The game must feel fun, responsive, and polished within an arcade scope - not a tech demo, but a finished playable loop from title screen to game over.;

\%problem No game exists yet; we need a single, self-contained, platform-independent implementation that anyone can run instantly without installing engines, package managers, or build tools.;

\%goal \{
  deliver a complete, immediately playable 2D retro action game;
  success criteria: runs by opening one file; smooth 60 FPS; tight controls; at least 4 distinct enemy behaviors; a full game-state loop (title -> play -> pause -> game over -> restart); visible score, lives, and health; satisfying audio-visual feedback;
\}

\%techs \{
  pure HTML5 + Canvas 2D + vanilla JavaScript (ES6+);
  Web Audio API for procedurally generated sound effects (no audio files);
  all graphics drawn programmatically or as embedded pixel data (no image files);
\}

\%mustnot \{
  no external libraries, frameworks, or CDNs;
  no build step, bundler, or package manager;
  no network requests of any kind;
  no platform-specific APIs (must run in any modern browser on any OS);
  no external asset files (images, audio, fonts);
  do not hallucinate APIs; use only standard web platform features;
\}

\%aware \{
  the implementation must be a single self-contained .html file;
  localStorage may be unavailable in some sandboxes - guard high-score persistence with try/catch and degrade gracefully to in-memory storage;
  keyboard layouts differ; support both WASD and arrow keys;
\}

\%risk variable monitor refresh rates can break physics; mitigate with a fixed-timestep update loop decoupled from requestAnimationFrame rendering;

\%risk unbounded entity spawning can degrade performance; mitigate with object pooling or hard caps on particles and projectiles;

\%label penguin-blizzard;

\%rule \%if <a new feature conflicts with the single-file, zero-dependency constraint> \{drop or simplify the feature; the constraint always wins;\};

\%rule \%if <player health reaches 0> \{lose one life; \%if <lives remain> \{respawn with brief invincibility;\} \%else \{transition to game-over state;\}\};

\%rule \%if <player collides with an enemy while not invincible> \{take damage; apply knockback; flash sprite; grant ~1.5s invincibility frames;\};

\%goldrule01 output should be grounded and mapped based on evidence; no hallucination;

\%reqfunc PLAYER: penguin character with idle, walk/waddle, jump, and attack animations rendered as programmatic pixel art; movement: left/right run with acceleration and friction; variable-height jump (hold = higher); optional belly-slide dash; attack: throw snowball projectiles with a fire-rate limit;

\%reqfunc ENEMY-WALKER: ground patroller that walks between edges or obstacles, turning at boundaries; damages player on contact;

\%reqfunc ENEMY-FLYER: airborne enemy following a sine-wave or swooping path; periodically dives toward the player;

\%reqfunc ENEMY-SHOOTER: stationary or slow-moving enemy that aims and fires projectiles at the player on a cooldown; projectiles are dodgeable (visible travel time);

\%reqfunc ENEMY-CHARGER: enemy that idles until the player enters its line of sight, telegraphs (brief wind-up animation), then charges fast in a straight line; stuns itself briefly on hitting a wall;

\%reqfunc COMBAT: snowballs destroy enemies (1-3 hits depending on type); destroyed enemies award score and burst into particles; optional stomp-on-head kill for walkers;

\%reqfunc LEVEL: side-scrolling level built from a tile map (platforms, hazards like ice spikes, gaps); camera follows the player with smoothing and level-bound clamping; at least 3 waves or sections of escalating difficulty, then loop with increased speed/spawn rate for endless arcade play;

\%reqfunc PICKUPS: collectible fish (score), hearts (restore health), and at least 2 power-ups (e.g., rapid-fire snowballs, temporary shield) with on-screen timers;

\%reqfunc HUD: score, high score, lives (icons), health (hearts or bar), active power-up indicator;

\%reqfunc STATES: title screen with controls explained; playing; pause (P or Esc); game over with final score and restart prompt; all transitions handled by a simple state machine;

\%reqfunc INPUT: keyboard (arrows + WASD, space/Z to jump, X/J to shoot, P/Esc to pause); prevent default browser scrolling on game keys;

\%reqfunc AUDIO: procedurally synthesized retro SFX via Web Audio (jump, shoot, hit, enemy death, pickup, game over); simple mute toggle (M); initialize audio context on first user input to satisfy autoplay policies;

\%reqfunc FEEDBACK: particle effects (snow puffs, enemy bursts), screen shake on damage, hit-flash, floating score popups, subtle ambient snowfall in the background;

\%reqnonfunc PERFORMANCE: stable 60 FPS on modest hardware; fixed-timestep logic (e.g., 60 Hz) with render interpolation or accumulator pattern;

\%reqnonfunc PORTABILITY: runs in any modern desktop browser by double-clicking the file; no server required;

\%reqnonfunc CODE-QUALITY: organized into clear classes/modules within the single file (Game, Player, Enemy subclasses, Particle, Level, Input, Audio, UI); constants grouped in a single tunable CONFIG object; commented sections;

\%reqnonfunc RELIABILITY: no uncaught exceptions during normal play; entity arrays cleaned each frame; pause fully freezes simulation;

\%reqnonfunc READABILITY: pixel-art style with a cohesive limited palette (icy blues, whites, dark night-sky background); crisp scaling (imageSmoothingEnabled = false);

\%should responsive canvas that scales to fit the window while preserving aspect ratio and pixel crispness;

\%should high-score persistence via guarded localStorage;

\%should difficulty curve: enemy speed, count, and shooter fire rate scale gradually with score or wave number;

\%could simple boss penguin-seal encounter after wave 3;

\%could basic touch controls (on-screen buttons) for mobile browsers;

\%could parallax background layers (distant icebergs, aurora);

\%optional gamepad API support if it adds no complexity to the input layer;

\%example enemy behavior contract: a CHARGER at rest plays an idle bob; when the player crosses its horizontal sightline within ~250px, it flashes for 400ms (telegraph), then accelerates to 3x walker speed until it hits a wall (1s self-stun, vulnerable) or leaves the screen bounds;

\%example feel target: jumping should resemble classic 8/16-bit platformers - fast rise, slightly faster fall (higher downward gravity), ~80ms coyote time, ~100ms jump input buffer;

\%plan produce the full game in one pass as a single self-contained HTML file, structured so a developer can tune gameplay via the CONFIG object without touching logic;

\%1 define CONFIG constants (physics, speeds, spawn rates, palette) and the fixed-timestep game loop skeleton;

\%2 implement input handling (keyboard mapping, key-state tracking, pause toggle);

\%3 implement the Player class: movement physics, jump with coyote time and input buffering, animation states, snowball firing, damage/invincibility;

\%4 implement the tile-based level, collision resolution (AABB vs tiles), hazards, and smoothed camera;

\%5 implement the four enemy types as a shared Enemy base class with per-type state machines (walker, flyer, shooter, charger) plus enemy projectiles;

\%6 implement combat resolution, pickups, power-ups, scoring, lives, and wave/difficulty progression;

\%7 implement HUD, title/pause/game-over screens, and the global state machine;

\%8 implement Web Audio SFX synthesis, particles, screen shake, and ambient snowfall polish;

\%9 final integration pass: object pooling/caps, error guards, code comments, and a manual test checklist (all states reachable, all enemies behave per spec, no console errors);

\%showplan before writing code, present a brief implementation plan and the planned CONFIG structure;

\%trace after delivering the code, summarize which requirement (\%reqfunc/\%reqnonfunc IDs above) maps to which section of the code;

\%diagram <game architecture: game loop, state machine, entity hierarchy, input/audio/render subsystems> \{type: flowchart;\};

\%diagram <player and enemy interaction flow: spawn -> detect -> telegraph -> attack -> damage/score> \{type: sequenceDiagram;\};

\%in none; everything is generated from this specification;

\%format markdown for explanations; a single complete HTML code block for the game itself;

\%out \{
  1. the execution plan (per \%showplan);
  2. the two diagrams (per \%diagram);
  3. one complete, runnable, self-contained .html file containing the entire game;
  4. a short requirements-to-code trace table (per \%trace);
  5. brief play instructions and tuning notes for the CONFIG object;
\}

\clearpage
\section{Appendix. Reverse Prompt: A Simplified Specification}
\label{app:penguin-prompt-b}


You are an expert game developer. Build a complete, polished, platform-independent 2D retro action game called \textbf{Penguin Blitz} according to the specification below. The game must be fully playable, fun, and self-contained.

\textbf{Hard Constraints (non-negotiable)}

\begin{itemize}
    \item \textbf{Zero external dependencies.} No third-party libraries, frameworks, engines, package managers, or downloaded assets. Use only what ships with the chosen runtime.
    \item \textbf{Platform independence.} The game must run unmodified on Windows, macOS, and Linux. Recommended approach: a single HTML file using vanilla JavaScript and the HTML5 Canvas 2D API plus the Web Audio API (runs in any modern browser on any OS). If another stack is chosen, it must use only the language's standard library.
    \item \textbf{No external assets.} All graphics must be drawn procedurally (rectangles, pixel-art drawn from code/arrays, shapes) and all sound effects/music must be synthesized at runtime.
    \item \textbf{Single-file delivery preferred} (e.g., one \texttt{index.html}). If multiple files are unavoidable, no build step may be required.
    \item \textbf{60 FPS target} with a fixed-timestep or delta-time game loop that behaves identically across machines.
\end{itemize}

\textbf{\textbf{Visual} Style}

\begin{itemize}
    \item Retro 8/16-bit pixel-art aesthetic: chunky pixels, limited palette (suggest 12--16 colors), scanline or CRT-style optional toggle.
    \item Integer pixel scaling with crisp rendering (disable image smoothing).
    \item Arctic/ice theme: snow, ice platforms, frozen caves, aurora background with simple parallax scrolling (2--3 layers).
    \item Readable silhouettes: player and each enemy type must be instantly distinguishable by shape and color.
\end{itemize}

\textbf{Player Character --- The Penguin}

\begin{itemize}
    \item A plucky pixel-art penguin, roughly $16 \times 16$ or $24 \times 24$ logical pixels, with idle, walk, jump, slide, and hurt animation frames (procedurally defined sprite frames are fine).
    \item \textbf{Movement \& abilities:}
    \begin{itemize}
        \item Left/right run with slight acceleration and friction (ice surfaces are extra slippery).
        \item Jump with variable height (hold = higher), coyote time ($\sim$100\,ms), and jump buffering ($\sim$120\,ms).
        \item \textbf{Belly slide}: a fast dash along the ground that damages enemies on contact and passes under low obstacles; short cooldown.
        \item \textbf{Snowball throw}: primary ranged attack; arcing or straight projectile with a fire-rate limit and satisfying impact effect.
    \end{itemize}
    \item \textbf{Health \& lives:} 3 hit points shown as hearts, 3 lives, brief invincibility frames with sprite flicker after taking damage, knockback on hit.
    \item Responsive controls: keyboard (arrows/WASD + Z/X or J/K) and remappable or at least documented. Add touch controls (on-screen buttons) if running on mobile browsers.
\end{itemize}

\textbf{Enemy Types}

Each enemy needs distinct movement, attack behavior, health, point value, a small death animation (puff/particles), and fair telegraphing.

\begin{enumerate}
    \item \textbf{Walker} --- patrols platforms, turns at edges/walls; damages on contact; 1~HP; can be defeated by jump-stomp, slide, or snowball.
    \item \textbf{Flyer} --- hovers in sine-wave or circular patterns; periodically swoops toward the player; 1~HP; immune to slide (must be shot or stomped).
    \item \textbf{Shooter} --- stationary or slow-moving turret-like enemy (e.g., a walrus with an ice cannon); fires projectiles at intervals or aimed at the player; projectiles must be visible and dodgeable; 2~HP.
    \item \textbf{Charger} --- idles until the player enters its line of sight, telegraphs (shake/flash $\sim$0.5\,s), then charges fast in a straight line until hitting a wall (briefly stunned, vulnerable); 2~HP.
\end{enumerate}

Include simple spawning/placement via level data, and scale enemy density with difficulty.

\textbf{Level Design \& Progression}

\begin{itemize}
    \item At least \textbf{5 handcrafted levels} defined as tile maps in code (ASCII or numeric arrays), each introducing a new mechanic or enemy, plus a final challenge level.
    \item Tile types: solid ground, ice (slippery), one-way platforms, spikes/hazards, breakable ice blocks, springs, checkpoints, level-exit flag/igloo.
    \item Smooth camera that follows the player with a dead-zone and look-ahead; camera clamped to level bounds.
    \item Optional: a simple boss at the end (e.g., a giant polar bear combining shooter + charger behavior with 3 phases).
\end{itemize}

\textbf{Pickups \& Power-ups}

\begin{itemize}
    \item \textbf{Fish} = coins/score; collecting 100 grants an extra life.
    \item \textbf{Heart} = restores 1~HP.
    \item \textbf{Power-ups (timed, $\sim$10\,s):} rapid-fire snowballs, triple-shot spread, speed boost, temporary shield/invincibility star.
    \item Floating/bobbing animation, sparkle particles, distinct pickup sounds.
\end{itemize}

\textbf{Game Flow \& States}

Implement a clean state machine: \textbf{Title screen $\rightarrow$ (optional How-to-Play) $\rightarrow$ Gameplay $\rightarrow$ Pause $\rightarrow$ Level Complete $\rightarrow$ Game Over $\rightarrow$ Win screen/credits.}

\begin{itemize}
    \item Title screen with animated logo, ``Press Start'', and high-score display.
    \item Pause menu (resume, restart, mute, quit to title).
    \item Level-complete screen with score tally (time bonus, fish collected, enemies defeated).
    \item Game over with continue option; win screen with total score and play time.
\end{itemize}

\textbf{Scoring, Difficulty \& Persistence}

\begin{itemize}
    \item Points for enemies (per type), fish, time bonuses, and a combo multiplier for consecutive kills without taking damage.
    \item Three selectable difficulties (Easy/Normal/Hard) affecting enemy speed, density, and player HP.
    \item Persist high scores and settings locally (e.g., \texttt{localStorage} in the browser); degrade gracefully if storage is unavailable.
\end{itemize}

\textbf{Audio (fully synthesized)}

\begin{itemize}
    \item Web Audio API (or stdlib equivalent) for: jump, shoot, hit, enemy death, pickup, power-up, slide, checkpoint, level complete, game over.
    \item A simple looping chiptune background track per level group (square/triangle oscillators, basic note sequencer), with separate music/SFX volume controls and a mute toggle.
\end{itemize}

\textbf{Juice \& Polish}

\begin{itemize}
    \item Particle effects: snow puffs on landing, hit sparks, death explosions, ambient falling snow.
    \item Screen shake on heavy hits, hit-stop (2--4 frames) on enemy kills, flash effects on damage.
    \item Squash-and-stretch on jump/land; smooth UI transitions (fade in/out between states).
    \item HUD: hearts, lives, score, fish count, active power-up timer --- pixel-styled and unobtrusive.
\end{itemize}

\textbf{Technical Quality Requirements}

\begin{itemize}
    \item Clean, well-commented, modular code: separate concerns (input, physics, entities, rendering, audio, level data, state management), ideally via lightweight classes or modules within the single file.
    \item AABB collision detection with proper tile-based collision resolution (no tunneling at high speeds --- use swept checks or sub-stepping).
    \item An entity system that makes adding new enemy types trivial.
    \item No memory leaks: pool or properly remove dead entities and finished particles.
    \item Pause/resume must fully halt simulation and audio; the game must handle window blur/focus and canvas resize (maintain aspect ratio with letterboxing).
    \item Deterministic, frame-rate-independent physics.
\end{itemize}

\textbf{Acceptance Criteria (test before delivering)}

\begin{enumerate}
    \item Opens and runs from a double-clicked single file with no installation, no internet, no console errors.
    \item All four enemy types behave as specified and are each defeatable by at least one method.
    \item Player movement feels tight: coyote time, jump buffering, and variable jump verified.
    \item All 5+ levels are completable; checkpoints and lives work; game can be finished start to end.
    \item Audio plays (after first user interaction, per browser autoplay rules), and mute works.
    \item High score persists across page reloads.
    \item Stable 60 FPS on a mid-range laptop with 50+ entities on screen.
\end{enumerate}

Deliver the complete code, followed by a short README textbf covering controls, how to run it, and a brief architecture overview.


\clearpage
\section{Appendix. Reverse Prompt: Software Requirements Specification}
\label{app:penguin-prompt-srs}
%
%

\noindent\textbf{Version:} 1.0 \quad \textbf{Date:} June 12, 2026 \quad \textbf{Status:} Draft for AI-Assisted Implementation \\
\textbf{Intended Audience:} Third-party developers using AI coding assistants

\subsection{Introduction}

\textbf{Purpose}

This document specifies the complete functional and non-functional requirements for a 2D retro-style action game featuring a penguin protagonist. It is written to be handed directly to an AI coding assistant (or a developer working with one) as a self-contained implementation brief. Every requirement is intended to be unambiguous, testable, and implementable without further clarification.

\textbf{Scope}

The product is a single-player, side-scrolling 2D action game with retro (8/16-bit-inspired) aesthetics. The player controls a penguin who traverses horizontally scrolling levels, defeats four enemy archetypes (walkers, flyers, shooters, chargers), collects pickups, and progresses through levels of increasing difficulty.

The game must be \textbf{fully platform-independent with zero external dependencies}: no third-party libraries, frameworks, engines, asset files, or network resources. All graphics and audio must be generated procedurally in code.

\textbf{Definitions, Acronyms, and Abbreviations}
\begin{center}
\begin{tabular}{p{2.2cm} p{10.5cm}}
\toprule
\textbf{Term} & \textbf{Definition} \\
\midrule
SRS & Software Requirements Specification \\
FR & Functional Requirement \\
NFR & Non-Functional Requirement \\
HUD & Heads-Up Display (on-screen status overlay) \\
AABB & Axis-Aligned Bounding Box (rectangle collision) \\
Sprite & A 2D graphical object representing a game entity \\
Tile & A fixed-size square unit used to compose level geometry \\
i-frames & Invincibility frames; brief post-damage invulnerability \\
Game loop & The update--render cycle executed every frame \\
\bottomrule
\end{tabular}
\end{center}

\textbf{Reference Implementation Target}

To satisfy the zero-dependency constraint, the \textbf{recommended target is a single self-contained HTML file} using only the HTML5 Canvas 2D API, the Web Audio API, and vanilla JavaScript (ECMAScript 2020+). This runs unmodified on Windows, macOS, Linux, ChromeOS, iOS, and Android via any modern browser, with no installation, build step, or package manager.

An implementer may substitute another runtime (e.g., Python with only the standard library via \texttt{tkinter}) provided all requirements in this SRS are met, but the single-file HTML5 target is the reference assumption throughout.

\textbf{Document Conventions}

\begin{itemize}
  \item Requirements are numbered \texttt{FR-<AREA>-NNN} (functional) and \texttt{NFR-NNN} (non-functional).
  \item ``Shall'' denotes a mandatory requirement; ``should'' denotes a recommendation; ``may'' denotes an option.
  \item All tunable numeric values (speeds, health, timers) are stated as defaults and shall be defined as named constants in a single configuration section of the code so they can be tuned without code restructuring.
\end{itemize}

\subsection{Overall Description}

\textbf{Product Perspective}
A standalone, self-contained product. There are no servers, accounts, saved cloud data, analytics, advertisements, or external services. The entire game is one deliverable artifact (one file).

\textbf{Product Functions (Summary)}

\begin{enumerate}
  \item Run a fixed-timestep game loop with decoupled rendering.
  \item Render procedurally drawn retro graphics on a virtual low-resolution canvas.
  \item Accept keyboard input (primary), with optional gamepad and touch input.
  \item Simulate 2D platformer physics: gravity, jumping, ground/wall collision.
  \item Control a penguin player character with movement, jump, and two attacks.
  \item Spawn and simulate four enemy archetypes with distinct AI behaviors.
  \item Manage combat: hit detection, damage, knockback, invincibility frames, enemy death.
  \item Present at least three handcrafted levels plus a boss-free final gauntlet, with a defined level format permitting easy addition of new levels.
  \item Track score, lives, health, and collectibles; display them on a HUD.
  \item Manage game states: title, playing, paused, level complete, game over, victory.
  \item Generate all sound effects and music procedurally via synthesized audio.
  \item Persist the local high score where the platform allows (e.g., \texttt{localStorage}), degrading gracefully if unavailable.
\end{enumerate}

\textbf{User Characteristics}

Casual players aged roughly 8+, familiar with classic platformer conventions (arrow keys / WASD, jump on a button). No tutorial text beyond a single controls screen is required.

\textbf{Constraints}

\begin{description}
  \item[C-1:] Zero external dependencies. No libraries, frameworks, fonts, images, audio files, or network requests of any kind.
  \item[C-2:] Single-file deliverable. All HTML, CSS, JavaScript, and data (level definitions, palettes) reside in one file.
  \item[C-3:] The game shall function offline, opened directly from the local filesystem (\texttt{file://}).
  \item[C-4:] No use of \texttt{eval}, dynamic code loading, or browser-permission-gated APIs (camera, geolocation, etc.).
  \item[C-5:] Audio shall only start after the first user interaction (browser autoplay policy compliance).
\end{description}

\textbf{Assumptions}

\begin{itemize}
  \item A modern evergreen browser (released within the last 4 years) is available.
  \item A physical or virtual keyboard exists; touch controls are a fallback, not the primary design target.
\end{itemize}

\subsection{Functional Requirements}

\textbf{Game Loop and Timing (FR-LOOP)}

\begin{description}
  \item[FR-LOOP-001:] The game shall use a fixed-timestep simulation of 60 updates per second, with rendering driven by \texttt{requestAnimationFrame} and an accumulator to decouple simulation from display refresh rate.
  \item[FR-LOOP-002:] If the tab loses focus or frame delta exceeds 250\,ms, the accumulated time shall be clamped to prevent the ``spiral of death'' and physics tunneling.
  \item[FR-LOOP-003:] All gameplay speeds shall be expressed in units per second (converted per fixed step), never per frame, so behavior is identical on 60\,Hz and 144\,Hz displays.
\end{description}

\textbf{Rendering and Visual Style (FR-GFX)}

\begin{description}
  \item[FR-GFX-001:] The game shall render to a virtual canvas of $320 \times 180$ pixels, integer-scaled to the largest size fitting the window, with image smoothing disabled to preserve crisp pixels.
  \item[FR-GFX-002:] All sprites shall be drawn procedurally (canvas primitives or pixel-array stamping from in-code arrays). No external image assets.
  \item[FR-GFX-003:] The game shall use a fixed retro palette of at most 32 colors, defined as named constants in one place.
  \item[FR-GFX-004:] The camera shall follow the player horizontally with a smoothing factor (lerp $\approx 0.1$ per step), clamped to level bounds, with a small look-ahead ($\approx 24$\,px) in the direction of movement.
  \item[FR-GFX-005:] The background shall include at least two parallax layers (e.g., distant icebergs at $0.25\times$ scroll, near hills at $0.5\times$ scroll).
  \item[FR-GFX-006:] Entity animations (walk, jump, attack, hit, death) shall use frame-based procedural animation at 6--12 animation frames per second.
  \item[FR-GFX-007:] Particle effects shall be present for: enemy death (burst of 8--12 particles), player damage, snowball impact, pickup collection, and player slide (snow spray).
  \item[FR-GFX-008:] Screen shake ($\leq 3$\,px, $\leq 200$\,ms) shall trigger on player damage and charger wall impact.
\end{description}

\textbf{Input (FR-INP)}

\begin{description}
  \item[FR-INP-001:] Keyboard controls (remapping not required):
  \begin{center}
  \begin{tabular}{p{4cm} p{4cm} p{4cm}}
  \toprule
  \textbf{Action} & \textbf{Primary} & \textbf{Secondary} \\
  \midrule
  Move left/right & Arrow Left/Right & A / D \\
  Jump & Space & W / Arrow Up \\
  Throw snowball & X & J \\
  Belly slide & Z & K \\
  Pause & Esc & P \\
  Confirm (menus) & Enter & Space \\
  \bottomrule
  \end{tabular}
  \end{center}
  \item[FR-INP-002:] Input shall distinguish ``held'' from ``just pressed'' states; jump and attacks trigger on just-pressed only.
  \item[FR-INP-003:] A jump input buffer of 100\,ms and coyote time of 100\,ms shall be implemented (jump registers if pressed slightly before landing or slightly after leaving a ledge).
  \item[FR-INP-004 (optional):] Gamepad support via the Gamepad API may be added; if present, it shall map left stick/D-pad to movement, bottom face button to jump, left face button to snowball, right face button to slide.
  \item[FR-INP-005 (optional):] On touch devices, on-screen buttons (left, right, jump, snowball, slide) may be displayed; they shall not appear when a keyboard event has been received.
\end{description}

\textbf{Physics and Collision (FR-PHY)}

\begin{description}
  \item[FR-PHY-001:] Gravity shall be a constant downward acceleration (default 1500\,px/s\textsuperscript{2}) applied to all airborne dynamic entities, with terminal fall speed 420\,px/s.
  \item[FR-PHY-002:] Collision shall use AABB tests against a tile grid ($16 \times 16$\,px tiles), resolved axis-by-axis (horizontal then vertical) to prevent corner snagging.
  \item[FR-PHY-003:] Tile types shall include at minimum: empty, solid, one-way platform (solid only from above while falling), hazard spikes (damage on contact), and ice (reduced friction: 30\% of normal ground deceleration).
  \item[FR-PHY-004:] Entities shall never tunnel through solid tiles at maximum speeds (guaranteed by fixed timestep and per-axis resolution).
  \item[FR-PHY-005:] Falling below the bottom of the level shall cost the player one life and respawn them at the last checkpoint.
\end{description}

\textbf{Player Character --- Penguin (FR-PLY)}

\begin{description}
  \item[FR-PLY-001:] The player is an animated penguin sprite, approx. $14 \times 18$\,px, with distinct idle, walk, jump, fall, slide, throw, hurt, and death animations.
  \item[FR-PLY-002 (Movement):] Ground acceleration 1200\,px/s\textsuperscript{2} to a max run speed of 120\,px/s; ground deceleration 1400\,px/s\textsuperscript{2} (reduced on ice per FR-PHY-003); air control at 60\% of ground acceleration.
  \item[FR-PLY-003 (Jump):] Initial jump velocity $-330$\,px/s. Variable height: releasing jump while ascending shall cut vertical velocity by 50\%. Single jump only (no double jump in v1).
  \item[FR-PLY-004 (Snowball attack):] Pressing snowball throws a projectile (speed 240\,px/s, slight gravity 300\,px/s\textsuperscript{2}, range-limited to $\approx 1.5$\,s lifetime). Fire rate limited to one per 350\,ms. Max 3 player projectiles alive at once. Snowballs deal 1 damage and break on first contact with an enemy or solid tile.
  \item[FR-PLY-005 (Belly slide):] Pressing slide while grounded and moving initiates a slide: hitbox height reduces $\approx 40\%$, speed boosts to 190\,px/s decaying over 600\,ms, during which contact deals 1 damage to enemies and the player passes under low gaps. Slide on ice extends duration by 50\%. Cooldown 800\,ms.
  \item[FR-PLY-006 (Health and lives):] The player has 3 hit points (HP) and starts with 3 lives. Reaching 0 HP costs one life and respawns the player at the last checkpoint with full HP. Losing all lives triggers Game Over.
  \item[FR-PLY-007 (Damage response):] Taking damage applies knockback (away from the source, 160\,px/s horizontal, $-180$\,px/s vertical) and grants 1.2\,s of i-frames during which the sprite blinks and no damage can be taken.
  \item[FR-PLY-008 (Stomp):] Landing on an enemy's head while falling deals 1 damage to it and bounces the player ($-220$\,px/s), unless the enemy is specified as stomp-immune.
\end{description}

\textbf{Enemies (FR-ENM)}

\paragraph{Common Enemy Requirements}
\begin{description}
  \item[FR-ENM-001:] Each enemy type shall have: HP, contact damage (1 to the player unless stated), a movement behavior, distinct sprite/silhouette and color, hurt flash (white, 100\,ms), and a death sequence (particles + score popup) before removal.
  \item[FR-ENM-002:] Enemies shall be defined in level data by type and spawn position, and shall activate only when within 1.5 screen-widths of the camera (deactivate and reset when far off-screen).
  \item[FR-ENM-003:] Enemies shall not damage each other and shall not be affected by player knockback unless stated.
  \item[FR-ENM-004:] On death, each enemy awards score (see FR-SCO-001) and has a 10\% chance to drop a fish pickup (FR-PCK-002).
\end{description}

\paragraph{Walker --- ``Snow Slug'' (FR-ENM-W)}
\begin{description}
  \item[FR-ENM-W-001:] HP 1. Patrols horizontally at 40\,px/s.
  \item[FR-ENM-W-002:] Reverses direction on hitting a wall or reaching a ledge edge (shall not walk off platforms).
  \item[FR-ENM-W-003:] Vulnerable to stomp, snowball, and slide.
\end{description}

\paragraph{Flyer --- ``Frost Bat'' (FR-ENM-F)}
\begin{description}
  \item[FR-ENM-F-001:] HP 1. Ignores gravity. Flies along a sine-wave path: horizontal speed 60\,px/s, vertical amplitude 24\,px, period 2\,s, reversing at patrol bounds defined in level data (default $\pm 96$\,px from spawn).
  \item[FR-ENM-F-002:] If the player comes within 80\,px, the flyer shall swoop toward the player's position (speed 110\,px/s) for up to 1\,s, then return to its patrol altitude.
  \item[FR-ENM-F-003:] Vulnerable to snowball and stomp; slide cannot reach it unless it is low during a swoop.
\end{description}

\paragraph{Shooter --- ``Ice Lobber'' (FR-ENM-S)}
\begin{description}
  \item[FR-ENM-S-001:] HP 2. Stationary turret-style enemy anchored to the ground.
  \item[FR-ENM-S-002:] When the player is within 200\,px horizontally and roughly line-of-sight (no solid tile fully blocking), it fires an icicle projectile toward the player every 1.8\,s (projectile speed 150\,px/s, straight-line, 1 damage, destroyed on any collision). Enemy projectiles shall be visually distinct (pale blue, elongated) from player snowballs.
  \item[FR-ENM-S-003:] A 400\,ms wind-up animation (visible telegraph) shall precede every shot.
  \item[FR-ENM-S-004:] Vulnerable to snowball, stomp, and slide. Player snowballs and enemy icicles shall destroy each other on contact.
\end{description}

\paragraph{Charger --- ``Bull Seal'' (FR-ENM-C)}
\begin{description}
  \item[FR-ENM-C-001:] HP 2. Idles in place until the player is within 140\,px horizontally on the same approximate elevation ($\pm 32$\,px).
  \item[FR-ENM-C-002:] After a 500\,ms telegraph (rear-back animation + exclamation mark), it charges at 200\,px/s in the player's direction until it hits a wall (stunned 1.5\,s, vulnerable, triggers screen shake) or travels 320\,px (returns to idle).
  \item[FR-ENM-C-003:] Stomp-immune while charging (stomping a charging Bull Seal damages the player); stompable while idle or stunned. Always vulnerable to snowball. Slide vs. charge: both take 1 damage and knock back.
  \item[FR-ENM-C-004:] Contact damage during a charge is 1 HP with $1.5\times$ knockback.
\end{description}

\textbf{Pickups and Power-ups (FR-PCK)}

\begin{description}
  \item[FR-PCK-001 (Coins/Snowflakes):] Placed in level data; collecting one awards 50 points. Collecting 100 snowflakes grants an extra life (counter resets).
  \item[FR-PCK-002 (Fish):] Restores 1 HP (capped at max). Dropped by enemies (10\%) and placed sparsely in levels.
  \item[FR-PCK-003 (Golden Fish):] Rare placed pickup; grants temporary rapid-fire (snowball cooldown 120\,ms) for 8\,s, indicated by a HUD timer and player sprite tint.
  \item[FR-PCK-004 (1-Up Egg):] Rare placed pickup; grants one extra life.
  \item[FR-PCK-005:] All pickups shall bob gently (sine, $\pm 2$\,px) and emit a small particle burst plus a distinct sound when collected.
\end{description}

\textbf{Levels and Progression (FR-LVL)}

\begin{description}
  \item[FR-LVL-001:] Levels shall be authored as ASCII string arrays embedded in the code, one character per tile (e.g., \texttt{\#} solid, \texttt{-} one-way, \texttt{\^{}} spikes, \texttt{\textasciitilde} ice, \texttt{P} player spawn, \texttt{w}/\texttt{f}/\texttt{s}/\texttt{c} enemy spawns, \texttt{*} snowflake, \texttt{F} fish, \texttt{G} goal flag, \texttt{K} checkpoint). The legend shall be documented in code comments.
  \item[FR-LVL-002:] The game shall ship with at least 3 levels of increasing difficulty: L1 introduces movement + walkers; L2 adds flyers + shooters + ice tiles; L3 adds chargers + spike gauntlets + combined encounters.
  \item[FR-LVL-003:] Each level shall contain at least one checkpoint (\texttt{K}); crossing it sets the respawn point (visual state change on the checkpoint).
  \item[FR-LVL-004:] Touching the goal flag completes the level: a short jingle plays, a ``Level Complete'' summary screen shows score, snowflakes collected, and time, then advances to the next level.
  \item[FR-LVL-005:] Completing the final level triggers the Victory state with total score and a ``Thanks for playing!'' screen.
  \item[FR-LVL-006:] Minimum level length: 60 tiles (L1) growing to 120+ tiles (L3).
\end{description}

\textbf{Scoring (FR-SCO)}

\begin{description}
  \item[FR-SCO-001:] Score events: snowflake 50; fish 100; enemy kills --- walker 100, flyer 150, shooter 200, charger 250; level-complete time bonus $= \max(0, (180 - \mathrm{seconds}) \times 10)$.
  \item[FR-SCO-002:] Floating score popups shall appear at the point of each award, rising and fading over $\approx 0.7$\,s.
  \item[FR-SCO-003:] The highest total score shall persist locally (e.g., \texttt{localStorage} key \texttt{frostbite\_highscore}). If storage is unavailable (private mode, \texttt{file://} restrictions), the game shall continue without persistence and without errors.
\end{description}

\textbf{Game States and UI (FR-UI)}

\begin{description}
  \item[FR-UI-001:] State machine with states: TITLE, CONTROLS, PLAYING, PAUSED, LEVEL\_COMPLETE, GAME\_OVER, VICTORY. All transitions shall be explicit; no gameplay simulation occurs outside PLAYING.
  \item[FR-UI-002 (Title):] Game title, ``Press Enter to Start'', ``C for Controls'', and high score. Animated background permitted.
  \item[FR-UI-003 (HUD, during play):] Hearts for HP (3 max), lives counter (penguin icon $\times$ N), score, snowflake count, current level name, and active power-up timer. HUD elements shall be drawn with the in-code pixel font or canvas text styled monospace; HUD shall never obstruct the center of the screen.
  \item[FR-UI-004 (Pause):] Esc toggles pause; the screen dims, ``PAUSED'' displays, with options Resume and Quit to Title. Audio shall be suspended while paused.
  \item[FR-UI-005 (Game Over):] Shows final score and high-score notification if beaten; Enter returns to Title.
  \item[FR-UI-006:] All text shall be readable at the smallest supported window size ($\geq 640 \times 360$ CSS pixels).
\end{description}

\textbf{Audio (FR-AUD)}

\begin{description}
  \item[FR-AUD-001:] All audio shall be synthesized at runtime via the Web Audio API (oscillators + noise buffers + gain envelopes). No audio files.
  \item[FR-AUD-002:] Required sound effects (distinct, $\leq 0.5$\,s): jump, land, snowball throw, snowball impact, slide, player hurt, player death, enemy hit, enemy death, pickup, checkpoint, level complete jingle, menu confirm, charger telegraph, shooter fire.
  \item[FR-AUD-003:] A looping chiptune-style background track (simple 2--4 channel pattern sequencer: square lead, triangle bass, noise percussion) shall play during gameplay, with a different tempo/key per level. Music volume shall sit below SFX.
  \item[FR-AUD-004:] A master mute toggle (M key) shall be available at all times and reflected by a HUD icon. Audio context shall initialize only after first user input (per C-5).
\end{description}

\subsection{Non-Functional Requirements}

\begin{description}
  \item[NFR-001 (Platform independence):] The deliverable shall run unmodified on current Chrome, Firefox, Safari, and Edge on desktop and mobile, opened from \texttt{file://} or any static host, with zero installation.
  \item[NFR-002 (Dependencies):] Zero third-party code or assets. Only standard web platform APIs (Canvas 2D, Web Audio, Gamepad, localStorage, requestAnimationFrame).
  \item[NFR-003 (Performance):] Stable 60 FPS rendering with $\leq 4$\,ms average simulation time per frame on a 2018-era mid-range laptop; no per-frame allocations in hot loops where avoidable (object pooling for projectiles and particles).
  \item[NFR-004 (Size):] The single file should remain under 200\,KB uncompressed.
  \item[NFR-005 (Code quality):] Code shall be organized in clearly delimited sections (CONFIG, AUDIO, INPUT, PHYSICS, ENTITIES, LEVELS, STATES, RENDER, MAIN), with all tunable constants in CONFIG, and concise comments on every non-obvious algorithm (collision resolution, fixed timestep, audio synthesis).
  \item[NFR-006 (Robustness):] No uncaught exceptions during any state transition, window resize, rapid input mashing, or storage failure. Resizing the window at any time shall rescale the canvas without breaking gameplay.
  \item[NFR-007 (Determinism):] Given identical inputs, simulation behavior shall be frame-rate independent (consequence of FR-LOOP-001/003).
  \item[NFR-008 (Accessibility, minimum):] Color choices shall keep enemies and hazards distinguishable by shape/silhouette, not color alone; screen shake shall be subtle and brief.
\end{description}

\subsection{Acceptance Criteria (Verification Checklist)}

\begin{enumerate}
  \item Opening the single file in each major browser shows the title screen; Enter starts Level 1.
  \item The penguin runs, jumps with variable height, coyote time, and jump buffering; slides under a 1-tile gap; throws at most 3 concurrent snowballs.
  \item Each enemy demonstrably exhibits its specified behavior, telegraphs, vulnerabilities, and immunities (notably: stomping a charging Bull Seal hurts the player; snowballs cancel icicles).
  \item Spikes, falling off the level, and enemy contact damage all reduce HP/lives correctly, with knockback and i-frames; checkpoints respawn correctly.
  \item All three levels are completable; the victory screen shows after Level 3; high score persists across reloads where storage permits.
  \item Pausing freezes simulation and audio; muting silences everything instantly.
  \item Disconnecting from the network and reopening the file changes nothing (full offline operation).
  \item Frame-rate test: gameplay speed is identical on a 60\,Hz and a 144\,Hz monitor.
  \item The file contains no \texttt{<script src>}, \texttt{fetch}, \texttt{import}, \texttt{@font-face} URL, or any other external reference.
\end{enumerate}

\subsection{Suggested Implementation Order (for the AI Coding Session)}

\begin{enumerate}
  \item Canvas setup, integer scaling, fixed-timestep loop.
  \item Input layer (just-pressed/held abstraction).
  \item Tile map parsing + rendering + AABB collision.
  \item Player movement, jump feel (buffer/coyote/variable height), camera.
  \item Snowball + slide attacks, projectile pooling.
  \item Enemy framework, then Walker $\rightarrow$ Flyer $\rightarrow$ Shooter $\rightarrow$ Charger.
  \item Damage system, i-frames, lives, checkpoints, pickups.
  \item Levels 1--3 authored in ASCII; goal/checkpoint logic.
  \item State machine + HUD + menus.
  \item Procedural audio (SFX first, then music sequencer).
  \item Particles, screen shake, polish, high-score persistence.
  \item Run the acceptance checklist (Subsection A.5).
\end{enumerate}


\end{document}